\documentclass[10pt,twocolumn,letterpaper]{article}

\usepackage[pagenumbers]{wacv} %
\usepackage{maths}

\def\be{\begin{equation}}
\def\ee{\end{equation}}
\def\bea{\begin{eqnarray}}
\def\eea{\end{eqnarray}}

\def\sect#1{Section~\ref{sec:#1}}

\def\mypar#1{\vspace{1mm}{\noindent\bf #1}\hspace{1mm}}

\makeatletter
\DeclareRobustCommand\onedot{\futurelet\@let@token\@onedot}
\def\@onedot{\ifx\@let@token.\else.\null\fi\xspace}
\def\eg{{e.g}\onedot, }

\def\ie{{i.e}\onedot, }

\usepackage{graphicx}
\usepackage{amsmath}
\usepackage{amssymb}
\usepackage{booktabs}
\usepackage{cancel}
\usepackage{xcolor}
\definecolor{cvprblue}{rgb}{0.21,0.49,0.74}
\usepackage[pagebackref,breaklinks,colorlinks,citecolor=cvprblue]{hyperref}
 \usepackage{multirow,multicol}
\usepackage{tabularx}
\usepackage[capitalize]{cleveref}
\usepackage{tikz}
\usepackage[breakable]{tcolorbox}
\usepackage{amssymb}%
\usepackage{pifont}%
\usepackage{enumitem}%
\usepackage{xcolor,rotating}

\crefname{section}{Sec.}{Secs.}
\Crefname{section}{Section}{Sections}
\Crefname{table}{Table}{Tables}
\crefname{table}{Tab.}{Tabs.}
\makeatletter
\DeclareRobustCommand\onedot{\futurelet\@let@token\@onedot}
\def\@onedot{\ifx\@let@token.\else.\null\fi\xspace}

\def\eg{\textit{e.g}\onedot} 
\def\ie{\textit{i.e}\onedot}

\makeatother

\newcommand{\methodname}[0]{SANE\xspace}

\newcommand{\drawsquare}[2]{
    \begin{tikzpicture}
        \def\squaresize{#1}
        
        \fill[#2] (0,0) rectangle ++(\squaresize,\squaresize);
    \end{tikzpicture}
}

\newcommand{\drawrec}[2]{
    \begin{tikzpicture}
        \def\squaresize{#1}
                
        \fill[#2] (0,0) rectangle ++(\squaresize,2pt);
    \end{tikzpicture}
}

\definecolor{basecolor}{HTML}{A9A9A9}
\definecolor{cOne}{HTML}{FF7F0E}
\definecolor{cTwo}{HTML}{17BECF}
\definecolor{cThree}{HTML}{9467BD}

\newcommand{\sqGen}[0]{\drawsquare{0.4cm}{basecolor}}
\newcommand{\sqOne}[0]{\drawsquare{0.4cm}{cOne}}
\newcommand{\sqTwo}[0]{\drawsquare{0.4cm}{cTwo}}
\newcommand{\sqThree}[0]{\drawsquare{0.4cm}{cThree}}

\newcommand{\lGen}[0]{\drawrec{2.8cm}{basecolor}}
\newcommand{\lOne}[0]{\drawrec{2.8cm}{cOne}}
\newcommand{\lTwo}[0]{\drawrec{2.8cm}{cTwo}}
\newcommand{\lThree}[0]{\drawrec{2.8cm}{cThree}}

\newcommand{\sqHeader}[0]{\sqGen~ & \sqGen~\sqOne & \sqGen~\sqOne~\sqTwo & \sqGen~\sqOne~\sqTwo~\sqThree~}

\newcommand{\lines}[0]{& \raisebox{10pt}{\lGen~ }& \raisebox{10pt}{\lOne~ }& \raisebox{10pt}{\lTwo~ }& \raisebox{10pt}{\lThree~ }}

\newcommand{\itext}[1]{\scriptsize{\texttt{#1}}}

\newcolumntype{Y}{>{\centering\arraybackslash}X}

\newcolumntype{H}{>{\setbox0=\hbox\bgroup}c<{\egroup}@{}}

\definecolor{KleinBlue}{HTML}{8a2be2}
\definecolor{OliveGreen}{rgb}{0,0.6,0}
\definecolor{shadecolor}{rgb}{0.8,0.8,0.8}
\definecolor{DarkRed}{rgb}{0.55, 0.0, 0.0}

\usepackage{xstring}
\usepackage{ifthen}

\newcommand{\diff}[1]{%
  \if\relax\detokenize{#1}\relax %
    \textcolor{DarkRed}{\small{(#1)}}%
  \else
    \IfBeginWith{#1}{+}{%
      \small{(\textcolor{OliveGreen}{#1})}%
    }{%
      \small{(\textcolor{DarkRed}{#1})}%
    }%
  \fi
}

\begin{document}

\title{Specify and Edit: Overcoming Ambiguity in Text-Based Image Editing}

\author{Ekaterina Iakovleva\textsuperscript{1}\thanks{Equal contribution. Order decided randomly.}
\quad
Fabio Pizzati\textsuperscript{2}$^*$
\quad Philip Torr\textsuperscript{2} \quad Stéphane Lathuilière\textsuperscript{1} \\
\textsuperscript{1}\,LTCI, Télécom-Paris, Institut Polytechnique de Paris\\
\textsuperscript{2}\,University of Oxford 
}

\maketitle

\begin{abstract}
Text-based editing diffusion models exhibit limited performance when the user's input instruction is ambiguous. To solve this problem, we propose \textit{Specify ANd Edit} (\methodname), a zero-shot inference pipeline for diffusion-based editing systems. We use a large language model (LLM) to decompose the input instruction into specific instructions, \ie well-defined interventions to apply to the input image to satisfy the user's request. We benefit from the LLM-derived instructions along the original one, thanks to a novel denoising guidance strategy specifically designed for the task. Our experiments with three baselines and on two datasets demonstrate the benefits of \methodname in all setups. Moreover, our pipeline improves the interpretability of editing models, and boosts the output diversity. We also demonstrate that our approach can be applied to any edit, whether ambiguous or not. Our code is public at \url{https://github.com/fabvio/SANE}.\looseness=-1\end{abstract}

\section{Introduction}\label{sec:intro}

Recent advances in text-to-image generation have attracted a lot of attention in the research community and beyond it. This success is primarily due to development of text-conditioned diffusion models~\cite{rombach2022high,saharia2022photorealistic,ramesh2022hierarchical,nichol2021glide,couairon2023zero} which allow to translate textual concepts, described in natural language in the form of text prompts, into semantically coherent visualizations. Besides image synthesis, text-conditioned diffusion models have demonstrated strong performance on the image editing task~\cite{kawar2023imagic,hertz2022prompt,brooks2023instructpix2pix,sheynin2024emu,hui2024hq,zhang2024magicbrush}, where users describe in plain text the editing instructions to apply to input images.

\begin{figure}[t]
    \centering
    \includegraphics[width=\linewidth]{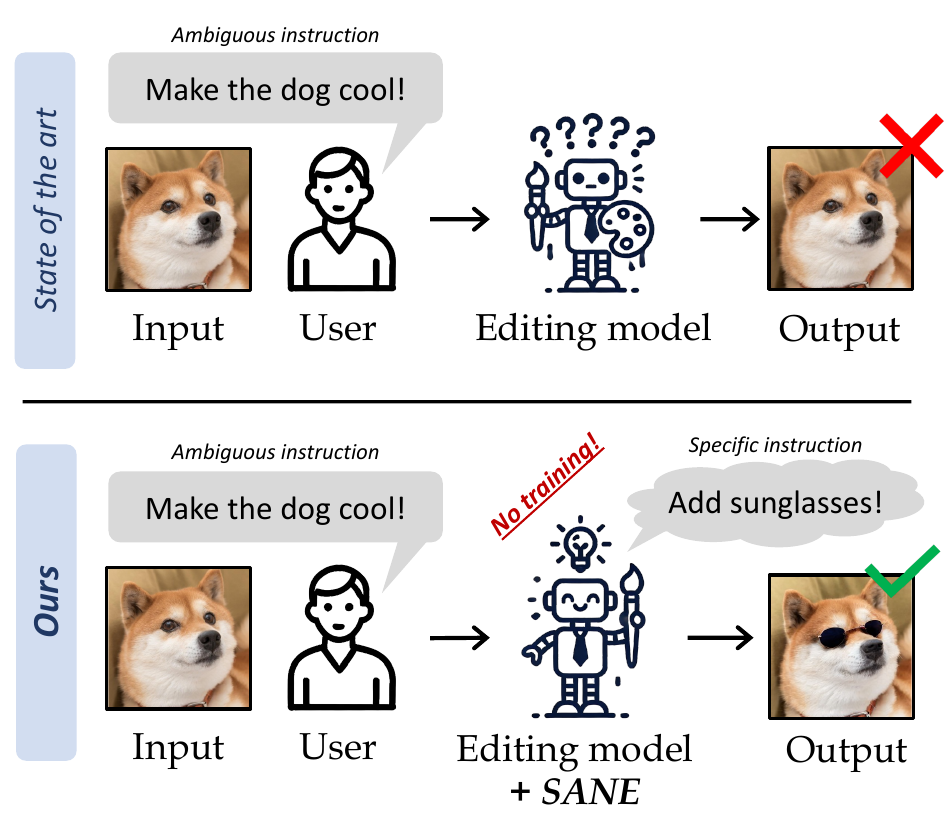}
    \caption{\textbf{Problem definition.} Abstract user instructions may lead existing editing diffusion models to failure (top). \methodname solves this problem by decomposing input instructions into specific ones, satisfying the user's request by integrating detailed edits in the editing process (bottom). \methodname is completely zero-shot, with no training required.\looseness=-1}
    \label{fig:teaser}
\end{figure}

Despite the remarkable success of text-conditioned editing methods, in this work, we start from the observation that these approaches usually fail to succesfully edit images when the edit prompt provided by the user is ambiguous.
To illustrate this, consider the example in Figure~\ref{fig:teaser}, which presents an editing task for a picture of a dog with the user instruction ``\textit{Make the dog cool.}" What does it mean for a dog to look \textit{cool}? Is there only one way for a dog to appear \textit{cool}?
To answer these questions, the editing model requires a nuanced contextual understanding. The same \textit{cool} adjective would suggest entirely different modifications if the subject were an inanimate object, like furniture, or a landscape. Moreover, there are multiple ways to make the dog look \textit{cool} (e.g., adding glasses, making it squint, changing the surroundings), all of which are equally valid.
 To address the multimodal nature of this task, editing models need reasoning and abstraction capabilities that current editing diffusion models lack~\cite{fu2023guiding}.

To address ambiguous input instructions, we propose our method \textit{Specify ANd Edit} (\methodname),  which leverages the reasoning capacities and the general knowledge of Large Language Models (LLMs)~~\cite{brown2020language,meta2024llama3,mistral2024mistral7b,anthropic2024claude3,anil2023gemini}. More precisely, \methodname breaks down the input edit into a series of specific instructions that, when applied together, transform the input into well-defined editing tasks. Through this process, known as \textit{specification}, we reverse the abstractions associated with input instructions. By incorporating various specific details, we effectively reduce the overall ambiguity of the input instructions.
After that, we condition a pre-trained editing diffusion model using both the specific and original ambiguous instructions. More precisely, we start with inferring individual noise estimations for each specific instruction using the chosen diffusion model. These noise estimations are then combined using a novel strategy described in \sect{method-instruction-combination}. Finally, the combined noise, along with the noise predicted from conditioning on the initial ambiguous instruction, is used in a modified classifier-free guidance~\cite{ho2022classifier}. This allows to preserve the fidelity to the original user indication, while guiding the process with the specific interventions. Among benefits on performance, this allows us to provide the specific instructions to the user at inference time, potentially raising the interpretability of the editing instruction. Notably, \methodname can be applied to an arbitrary pre-trained instruction-based diffusion model in a zero-shot manner.

In short, our contributions are: (i) We propose the first editing method designed specifically to address  ambiguous instructions. (ii) We introduce an LLM-based instruction decomposition pipeline, and a conditioning mechanism for editing diffusion models combining ambiguous and specific instructions, specifically designed for the task. Our approach requires no training. (iii) We perform extensive experiments on two datasets, with three state-of-the-art methods, outperforming all. We provide additional results on the properties of \methodname, and ablations.

\section{Related work}\label{sec:related}

\mypar{Diffusion-based Image Editing}
Diffusion models~\cite{sohl2015deep,ho2020denoising,rombach2022high,song2020denoising} have achieved remarkable results in generative image modeling by representing the generative process as a series of denoising steps. Conditioning these models on text has enabled controllable text-to-image synthesis~\cite{nichol2021glide,rombach2022high,ramesh2022hierarchical,saharia2022photorealistic},
as well as development of different diffusion-based editing systems~\cite{couairon2022diffedit,kawar2023imagic,nichol2021glide,meng2021sdedit,parmar2023zero,cao2023masactrl}. Some systems rely on image inversion technique~\cite{song2020denoising,dhariwal2021diffusion} to provide edited versions of an input scene~\cite{tumanyan2023plug, wallace2023edict, mokady2023null, miyake2023negative, cao2023masactrl, ju2023direct}. Although effective, these techniques are normally compute-intensive due to the inversion process. The seminal InstructPix2Pix~\cite{brooks2023instructpix2pix} solved this probelm by finetuning a diffusion model on instruction prompts, benefiting from synthetic pairs of images for training. Although many built on this result~\cite{sheynin2024emu,geng2024instructdiffusion,zhang2024magicbrush,hui2024hq,wang2023instructedit}, the impact of ambiguous instructions on instruction-based editing models is still not explored, motivating the proposed \methodname.\looseness=-1

\mypar{Large Language Models.}
LLMs~\cite{brown2020language,meta2024llama3,mistral2024mistral7b,anthropic2024claude3,anil2023gemini} are not only capable of human-like text completion, but are also successfully solving various reasoning tasks~\cite{anil2023palm,brown2020language,chowdhery2023palm,anil2023gemini,achiam2023gpt,romera2024mathematical}. This is achieved by applying various reasoning and prompting techniques, e.g. 
\textit{In-Context Learning} (ICL)~\cite{brown2020language}, where the model is given a few task examples in the form of demonstration~\cite{dong2022survey}. 
Another important direction of research is Visual-Language Models (VLMs)~\cite{clip,alayrac2022flamingo,li2023blip,huang2023tag2text,peng2023kosmos,liu2024visual} which aim to connect visual and language spaces. 
In this work, we use multimodal GPT-4o~\cite{openai2024gpt4o} to caption original images, and to decompose ambiguous instructions into sets of specific instructions using ICL and captions. 
There are several editing systems relevant to \methodname, which jointly finetune diffusion models and VLMs to enhance input instructions~\cite{fu2023guiding,huang2024smartedit}. 
While addressing a similar problem of commonsense reasoning for image editing, these models rely on implicit concept interpretations learned by VLMs. In contrast, \methodname is zero-shot and relies on explicit concept decomposition performed by GPT-4o. 

\mypar{Multi-instruction Editing}
There are several works that investigate multi-instruction scenarios for text-to-image generation and editing~\cite{guo2024focus,liu2022compositional,feng2024ranni,joseph2024iterative,fu2023guiding,huang2024smartedit,wang2023instructedit}. In image editing, the diffusion model is given a pre-defined set of instructions to follow. FoI~\cite{guo2024focus} extracts a region of interest for each of them, and restricts InstructPix2Pix~\cite{brooks2023instructpix2pix} to remain within the union of these regions. Instead of editing an image in one step, EMILIE~\cite{joseph2024iterative} applies InstructPix2Pix iteratively, one instruction at a time. To avoid artifacts, the authors propose to remain in the latent space of the diffusion model, and to decode only the last edited latent variable. 
CCA~\cite{hang2024cca} builds a multi-agent system that takes a composite instruction as input, then splits it into elementary instructions and iteratively applies them using an LLM, a VLM, and several editing diffusion models. 
In this work, we also consider sets of instructions, however, \methodname is the first to address instruction decomposition for ambiguous instructions.

\begin{figure*}[t]
    \centering
    \includegraphics[width=\linewidth]{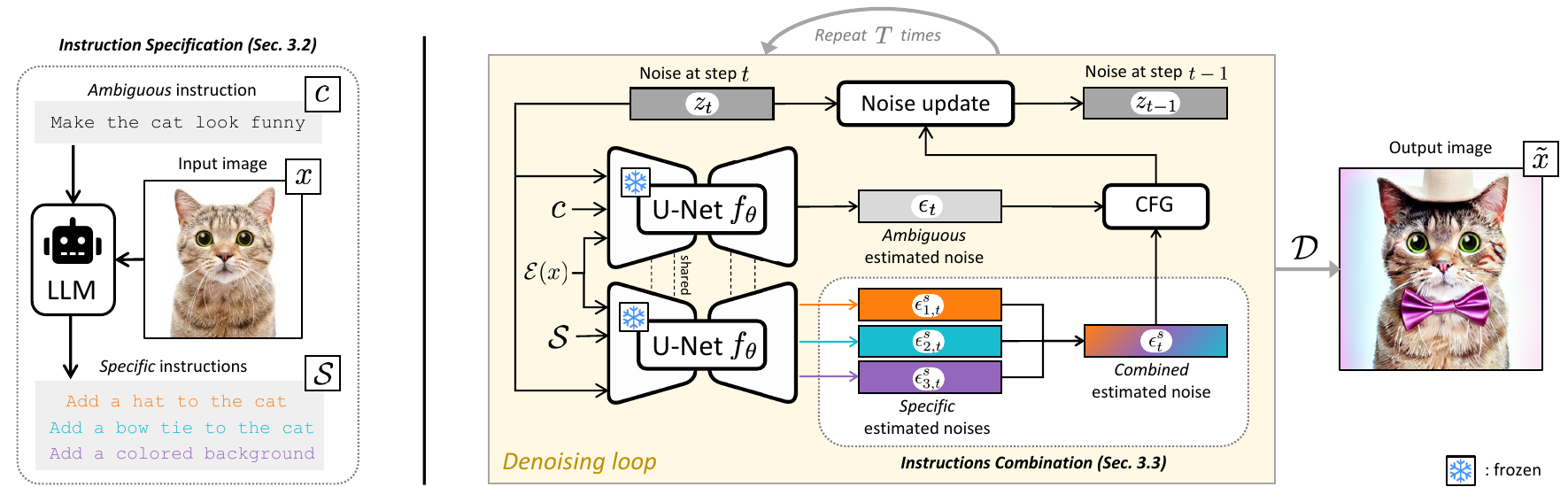}
    \caption{\textbf{\methodname inference pipeline.} We prompt an LLM to map an ambiguous input instruction $c$ to a set of specific instructions $\mathcal{S}$ (left). We provide a description of $x$ as context, in addition to $c$. Once $\mathcal{S}$ instructions are extracted, we use them in addition to $c$ in the denoising loop of an editing diffusion model (right). For each iteration, we estimate the noise in $z_{t}$ by conditioning the diffusion U-Net $f_\theta$ on all instructions. We them combine all specific instructions in a single noise estimation, that we later use in classifier-free guidance (CFG). We update the noise $z_t$ to $z_{t-1}$ following standard approaches. After $T$ iterations, we obtain the output image $\tilde{x}=\mathcal{D}(z_0)$.\looseness=-1}
    \label{fig:pipeline}
\end{figure*}

\section{Method}\label{sec:method}

This section introduces the \methodname pipeline. First, we present notations and preliminaries for diffusion-based image editing in Section~\ref{sec:preliminaries}. After that, we introduce our LLM-based instruction decomposition in Section~\ref{sec:method-instruction-decomposition}, and our 
novel instruction combination strategy in Section~\ref{sec:method-instruction-combination}.

\subsection{Background on Diffusion-based Image Editing}\label{sec:preliminaries}
We consider an instruction-based diffusion model such as InstructPix2Pix~\cite{brooks2023instructpix2pix}. The purpose of such models is to edit an input image $x$ while following a user-defined editing instruction $c$, in order to generate an edited image $\tilde{x}$. The edited image respects the input instruction while preserving the appearance of the original image. Following existing work~\cite{brooks2023instructpix2pix,zhang2024magicbrush,hui2024hq} we use a latent diffusion model~\cite{rombach2022high}, where the denoising operation is performed in the latent space of an autoencoder with encoder $\mathcal{E}$ and decoder $\mathcal{D}$. These models typically include a denoising U-Net $f_\theta$. 
At inference time, $\tilde{x}$ is produced by iterativly denoising a sampled Gaussian noise $z_T$ with $f_\theta$ over $T$ iterations. In other words, at step $t$, $f_\theta$ is used to estimate the noise $\epsilon_t$ in a corresponding noisy latent $z_t$. The latent $z_t$ is then updated to $z_{t-1}$, by removing the estimated noise $\epsilon_t$ and reintroducing Gaussian noise with lower intensity, following strategies from literature~\cite{ho2020denoising,song2020denoising,karras2022elucidating}. 
This is repeated for each $t \in [1, T]$, resulting in the final image $\tilde{x}=\mathcal{D}(z_0)$.

In instruction-based models, noise estimation is conditioned on the instruction $c$, which describes the desired changes, and on the input $x$, to enforce consistency with the original image. This noise is denoted as $\epsilon_t = f_\theta(z_t, \mathcal{E}(x), c)$. In practice, instruction-based diffusion models benefit from classifier-free guidance~\cite{ho2022classifier} to boost consistency towards the instruction and the input image~\cite{brooks2023instructpix2pix}. 
This means that for each $t$, $z_t$ is denoised using $\tilde{\epsilon}_t$, rather than $\epsilon_t$, with $\tilde{\epsilon}_t$ being a combination of three terms: unconditioned, conditioned on the image, and conditioned on the instruction $c$:
\begin{equation}\label{eq:cfgdef}
    \tilde{\epsilon}_t = \epsilon^\text{U}_t + \epsilon^\text{I}_t + \epsilon^\text{C}_t,
\end{equation}where
\begin{equation}\label{eq:cfg}
    \begin{aligned}
                &\epsilon^\text{U}_t = f_{\theta}(z_t, \varnothing, \varnothing),\\
        &\epsilon^\text{I}_t = w^\text{I}\cdot(f_\theta(z_t, \mathcal{E}(x), \varnothing) - f_\theta(z_t, \varnothing, \varnothing)),\\
        &\epsilon^\text{C}_t = w^\text{C}\cdot(\epsilon_t - f_\theta(z_t, \mathcal{E}(x), \varnothing)).
    \end{aligned}
\end{equation}
In Equation~\eqref{eq:cfg}, $\varnothing$ indicates that the conditioning element is replaced with zeros, and $w^\text{I}$ and $ w^\text{C}$ are guidance strength parameters, controlling the conditioning strength on $x$ and $c$, respectively.

\subsection{Instruction Specification}\label{sec:method-instruction-decomposition}
We noticed that directly applying ambiguous input instructions as $c$ may lead to limited editing performance. This is due to the ambiguity of $c$ which can be represented by multiple editing interventions. Let us assume that $x$ represents a cat, while $c=\text{``\texttt{make the cat look funny}''}$ is a user instruction, as in Figure~\ref{fig:pipeline} (left). 
As mentioned in Section~\ref{sec:intro}, we aim to map $c$ with an LLM to a set of specific and interpretable instructions that would satisfy the user's request, e.g. map $c$ to  ``\texttt{add a hat to the cat}''.

Formally, we want to extract from $c$ a set of $N$ editing instructions $\mathcal{S} = \{s_i\}_{i=1}^{N}$, where each $s_i$ is a specific instruction describing one modification consistent with $c$, to apply to the input scene. 
As shown in Figure~\ref{fig:pipeline} (left), we prompt an LLM to map $c$ to $N$ specific instructions, providing a rich caption of $x$ as context. 
Our prompt contains general descriptions of ambiguous and specific modifications, and two in-context learning examples~\cite{brown2020language} of ambiguous input instructions associated with desired specific instructions.
Furthermore, we set additional restrictions on the content and formatting style of the output instructions, to preserve image consistency and to simplify parsing. Due to limited space, we report the full prompt in the appendix. 
Notably, obtained specific instructions $\mathcal{S}$ are available to the user during model inference, providing insights on how the input instruction $c$ is respected. This also allows \methodname to explicitly show how $c$ impacts the input image $x$, which increases interpretability of the image editing process.

\subsection{Instructions Combination}\label{sec:method-instruction-combination}

After extracting decomposition $\mathcal{S} = \{s_i\}_{i=1}^{N}$, we can use it 
to guide the image editing process. Our idea is to include $\mathcal{S}$ in the denoising procedure of $f_\theta$ along the original ambiguous instruction $c$. Intuitively, this 
constrains
the state of solutions satisfying the required editing $c$, allowing the model to focus on the selected specific interventions $\mathcal{S}$. On the other hand, including $c$ in the denoising process allows the diffusion model to synthesize complementary elements necessary for satisfying the user request, but not included in $\mathcal{S}$. Hence, we propose a combination strategy that aggregate $c$ and each $s_i \in \mathcal{S}$, balancing the influence of the specific instructions without losing consistency with $c$.

We start by conditioning the denoising process on each specific instruction, to isolate their effects. Hence, for each $s_i \in \mathcal{S}$ we extract the estimated noise $\epsilon_{i,t}^s$ at timestep $t$:
\begin{equation}
    \epsilon_{i, t}^s = f_\theta(z_t, \mathcal{E}(x), s_i), \quad \forall i \in [1, N].
\end{equation}
This results in the set of estimated noises $\{\epsilon_{i, t}^{s}\}_{i=1}^N$, one for each $s_i \in \mathcal{S}$. 
Next, we aim to combine this set of noise estimations into a single noise estimate that aggregates all specific instructions. Simple averaging would diminish the impact of specific instructions that affect particular regions by blending them with other noise estimates. Therefore, we propose an alternative aggregation scheme, assuming that each image region is predominantly affected by a single specific instruction. To identify the spatial locations where the instruction $s_i$ most significantly impacts the diffusion process, we calculate the absolute difference between the estimated noises in the set and the noise obtained with conditioning only on the input image. Formally, it can be written as:
\begin{equation}
        \Delta \epsilon_{i,t}^s = |\epsilon_{i,t}^s - f_\theta(z_t, \mathcal{E}(x), \varnothing)|.
\end{equation}
We then aggregate these in a mask $M_t$, capturing the index of the most significant element across $i$:
\begin{equation}
    M_t = \arg\max_{i} \Delta \epsilon_{i,t}^s
\end{equation}
Finally, to aggregate the noises $\epsilon_{i,t}^s$ based on the mask $M_t$, we use $M_t$ to select the most significant estimated noise for each spatial location. The aggregated noise $\bar{\epsilon}_t^s$ can be computed as follows:

\begin{equation}
\Bar{\epsilon}_t^s = \sum_{i} \mathbb{I}(M_t = i) \cdot \epsilon_{i,t}^s,\label{eq:masking}
\end{equation}
where $\mathbb{I}(M_t = i)$ is an indicator function that is 1 if $M_t$ equals $i$, and 0 otherwise. Similarly to existing literature~\cite{liu2022compositional}, we use classifier-free guidance~\cite{ho2022classifier} for instruction combination, redefining Equation~\eqref{eq:cfgdef} as:
\begin{equation}
  \tilde{\epsilon}_t = \epsilon_t^\text{U} + \epsilon^\text{I}_t + \epsilon^\text{C}_t + \epsilon^\text{S}_t,
\end{equation}
where the classifier-free guidance term for the specific instructions is given by:
\begin{equation}
    \epsilon^\text{S}_t = w^\text{s}(\bar{\epsilon}_t^s - f_\theta(z_t, \mathcal{E}(x), \varnothing)).
\end{equation}
\noindent The process is shown in Figure~\ref{fig:pipeline}, center. We apply this for every iteration $t \in [1, T]$.

\begin{table*}[t]
\begin{subtable}{0.68\linewidth}
\setlength{\tabcolsep}{0.01\linewidth}
\resizebox{\linewidth}{!}{
    \begin{tabular}{cl|c|ccc|ccc}
    &\multicolumn{1}{c}{}& \multicolumn{1}{c}{}& \multicolumn{3}{c}{\textit{EMU-Edit}} & \multicolumn{3}{c}{\textit{IP2P data}}\\
    \toprule
    &\textbf{Method} & \textbf{$N$} & $\text{\textbf{CLIP}}_\text{\textbf{d}}\uparrow$ & $\text{\textbf{CLIP}}_\text{\textbf{i}}\uparrow$ & $\text{\textbf{CLIP}}_\Delta\uparrow$ & $\text{\textbf{CLIP}}_\text{\textbf{d}}\uparrow$ & $\text{\textbf{CLIP}}_\text{\textbf{i}}\uparrow$ & $\text{\textbf{CLIP}}_\Delta\uparrow$\\\midrule    \multirow{4}{*}{\rotatebox{90}{\textit{IP2P}}} & Baseline & - & 0.2923 & \textbf{0.8810} & 0.1203 & 0.2903 & \textbf{0.9027} & 0.1724 \\\cmidrule{2-9}

    & \multirow{3}{*}{\methodname} & 1 & 0.2961 & 0.7779 & 0.1705 & 0.2886 & 0.8196 & 0.1999 \\
    &&  2 & 0.2962 & 0.7654 & 0.1785 & 0.2910 & 0.8107 & \textbf{0.2063}\\
    &&  3 & \textbf{0.2968} & 0.7531 & \textbf{0.1858} & \textbf{0.2935} & 0.8101 & 0.2057\\\midrule
    \multirow{4}{*}{\rotatebox{90}{\textit{MB}}}& Baseline & - & 0.2888 & 0.7858 & 0.1618 & 0.2855 & 0.7934 & \textbf{0.2028} \\\cmidrule{2-9}
    &\multirow{3}{*}{\methodname} &  1 & 0.2948 & 0.7994 & 0.1661 & 0.2848 & 0.7970 & 0.1992 \\
    &&  2 & 0.2952 & 0.8134 & \textbf{0.1669} & 0.2870 & 0.8068 & 0.1996\\
    &&  3 & \textbf{0.3006} & \textbf{0.8209} & 0.1655 & \textbf{0.2878} & \textbf{0.8056} & 0.1998\\\midrule
    \multirow{4}{*}{\rotatebox{90}{\textit{HQEdit}}}& Baseline & - & 0.2675 & 0.6501 & 0.1417 & 0.2785 & 0.7035 & 0.1848 \\\cmidrule{2-9}
    &\multirow{3}{*}{\methodname} &  1 & 0.2725 & 0.6546 & 0.1458 & 0.2734 & 0.6997 & 0.1759 \\
    &&  2 & 0.2789 & 0.6802 & 0.1433 & 0.2782 & 0.7057 & 0.1842 \\
    &&  3 & \textbf{0.2823} & \textbf{0.6870} & \textbf{0.1474} & \textbf{0.2818} & \textbf{0.7136} & \textbf{0.1855}\\

    \bottomrule
    \end{tabular}
}
\caption{Editing quality results}\label{tab:editing-quality}
\end{subtable}
\hfill
\begin{subfigure}{0.27\linewidth}
    \centering
    \includegraphics[width=\linewidth]{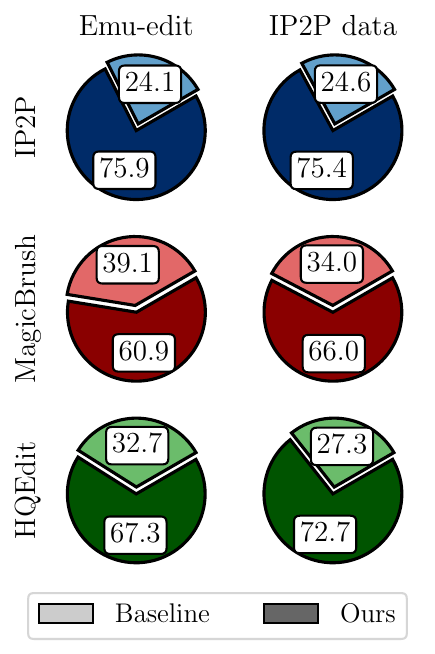}
    \caption{GPT evaluation}\label{tab:gpt-eval}
\end{subfigure}
\caption{\textbf{Quantitative comparison.} In~\subref{tab:editing-quality}, we compare against baselines by selecting different editing models and applying \methodname on top of them. We show results for $N=\{1, 2, 3\}$, evaluating $\text{CLIP}_\text{d}$, $\text{CLIP}_\text{i}$ and $\text{CLIP}_\Delta$ on all. \methodname consistently improve image quality performance across datasets and models. In~\subref{tab:gpt-eval}, we use GPT-4o to evaluate the quality of edited images, always outperforming baselines.}\label{tab:quant-clip}
\end{table*}

\begin{figure*}[t]
    \centering
    \setlength{\tabcolsep}{0pt} %
    \resizebox{\linewidth}{!}{
    \begin{tabularx}{\textwidth}{@{}H@{\hspace{0.01\textwidth}}Y@{\hspace{0.01\textwidth}}Y@{\hspace{0.01\textwidth}}YYY@{}}
    & & & \multicolumn{3}{c}{\methodname}\\
    & & Baseline & \small{$N=1$} & \small{$N=2$} & \small{$N=3$}\\
    \cmidrule(lr){3-3}\cmidrule(lr){4-6}
         & Original image & \sqHeader~ \\\midrule
         \raisebox{-5pt}{\rotatebox{90}{\textit{IP2P}}}
         & \includegraphics[width=0.19\textwidth, height=0.19\textwidth]{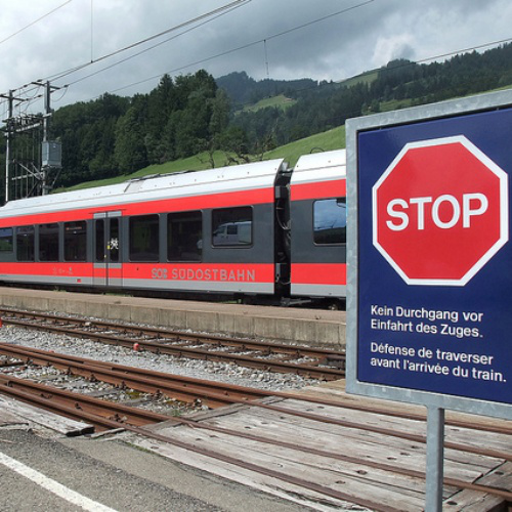} & 
         \includegraphics[width=0.19\textwidth, height=0.19\textwidth]{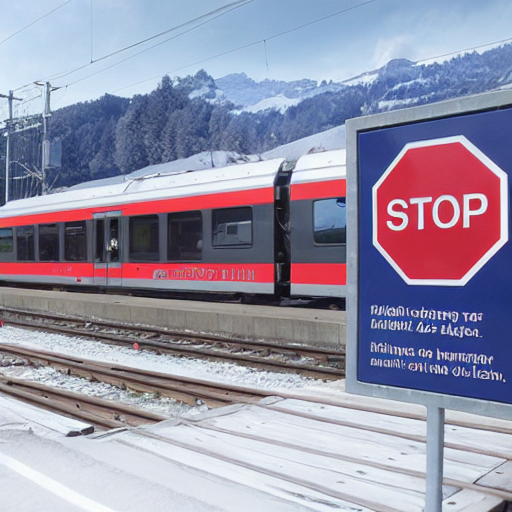} & 
         \includegraphics[width=0.19\textwidth, height=0.19\textwidth]{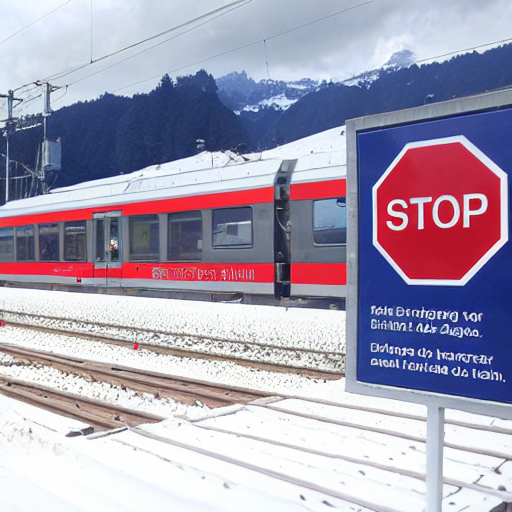} & 
         \includegraphics[width=0.19\textwidth, height=0.19\textwidth]{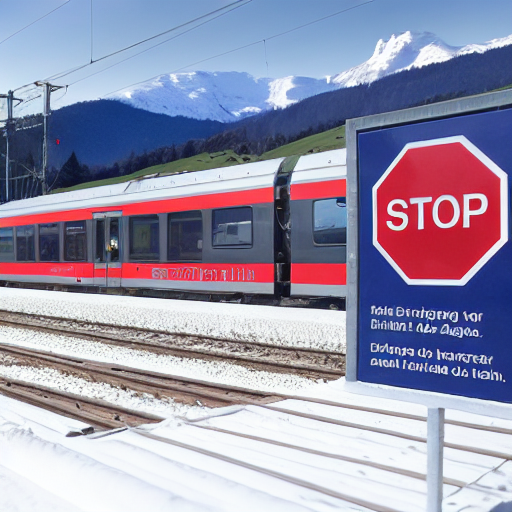} & 
         \includegraphics[width=0.19\textwidth, height=0.19\textwidth]{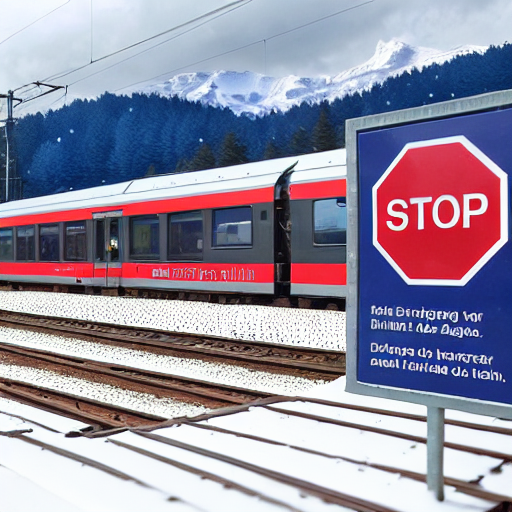} \\
         & Instructions~$\rightarrow$ & \itext{Change the scene into the alps in winter time} & \itext{Cover the ground with snow} & \itext{Add snowy mountain peaks in the background} & \itext{Add snow-covered pine trees along the track}  \\
         & \lines~ \\
         \multicolumn{6}{c}{\vspace{-20pt}}\\
         
         \midrule
         \raisebox{-20pt}{\rotatebox{90}{\textit{MB}}}
         & \includegraphics[width=0.19\textwidth, height=0.19\textwidth]{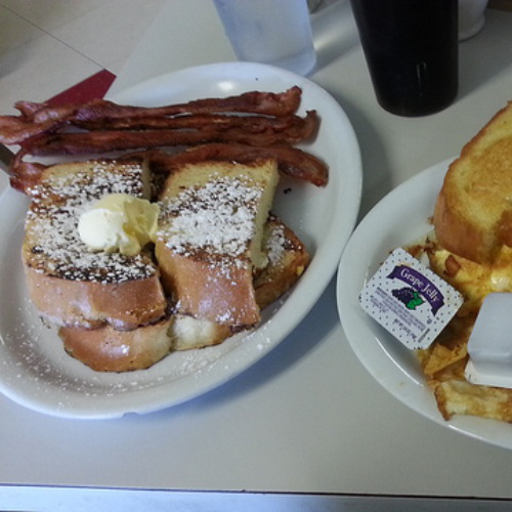} & 
         \includegraphics[width=0.19\textwidth, height=0.19\textwidth]{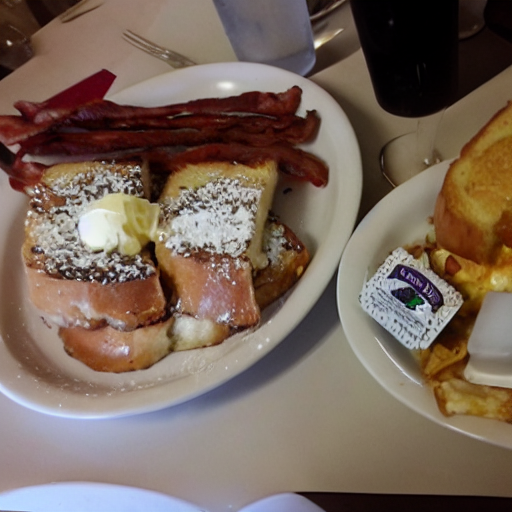} & 
         \includegraphics[width=0.19\textwidth, height=0.19\textwidth]{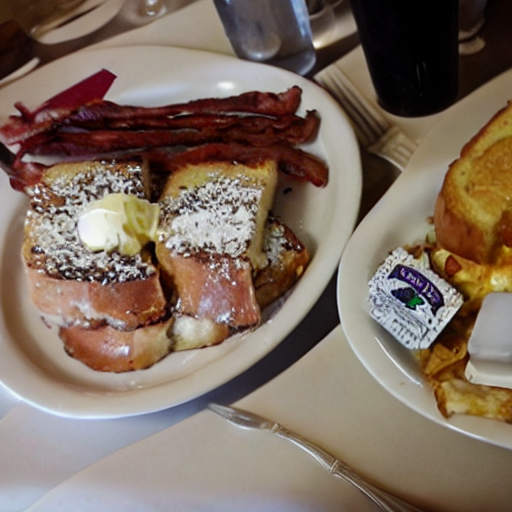} & 
         \includegraphics[width=0.19\textwidth, height=0.19\textwidth]{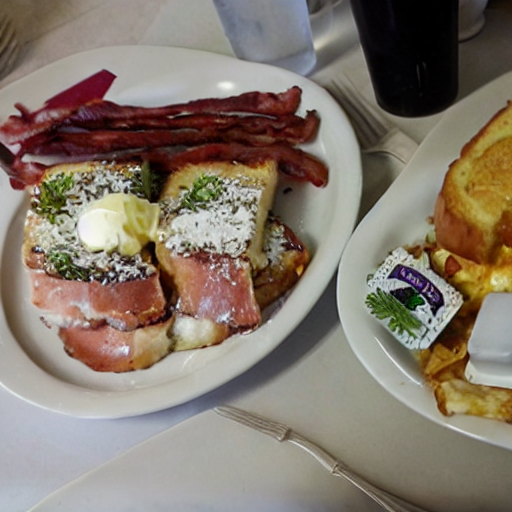} & 
         \includegraphics[width=0.19\textwidth, height=0.19\textwidth]{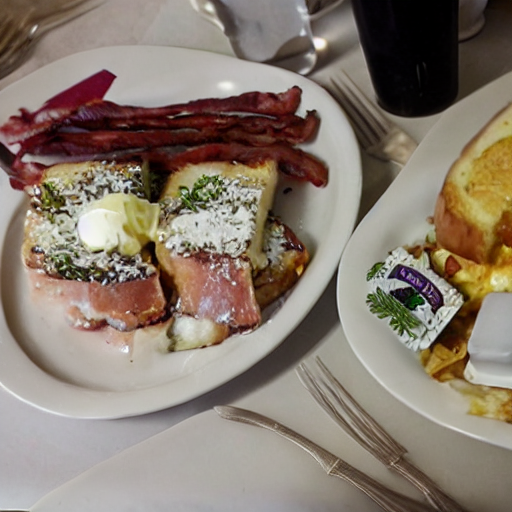} \\
         & Instructions~$\rightarrow$ & \itext{Make the photo seem like it was taken in a fancy restaurant} & \itext{Replace the plate with a gourmet presentation plate} & \itext{Add a garnish of fresh herbs to the bacon and eggs} & \itext{Add elegant silverware next to the plate}  \\
         & \lines~ \\
         \multicolumn{6}{c}{\vspace{-20pt}}\\
         
         \midrule

         \raisebox{-10pt}{\rotatebox{90}{\textit{HQEdit}}}
         & \includegraphics[width=0.19\textwidth, height=0.19\textwidth]{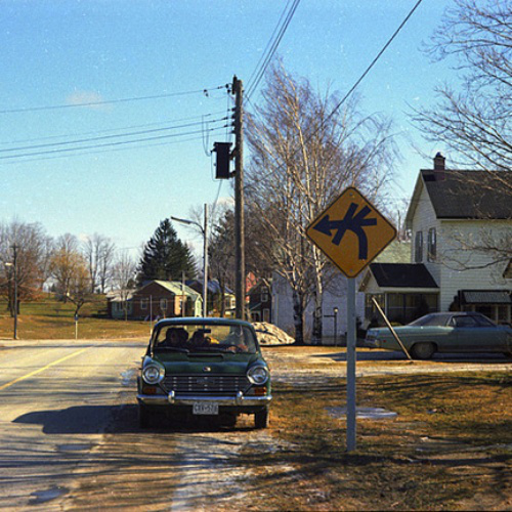} & 
         \includegraphics[width=0.19\textwidth, height=0.19\textwidth]{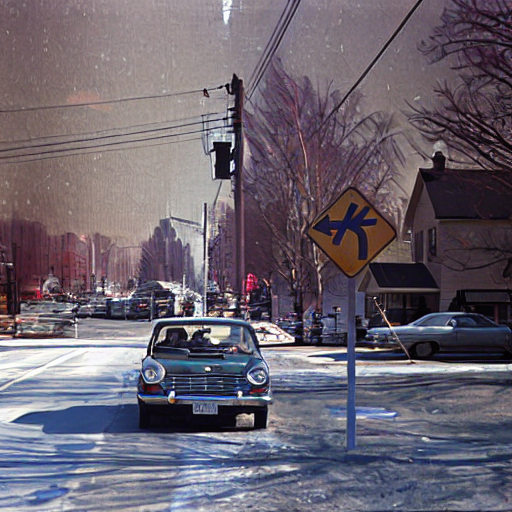} & 
         \includegraphics[width=0.19\textwidth, height=0.19\textwidth]{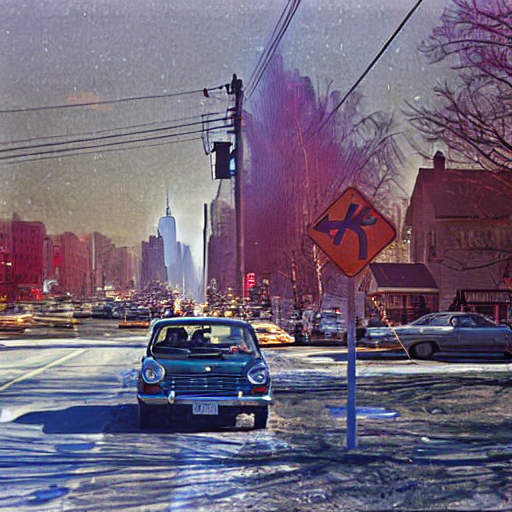} & 
         \includegraphics[width=0.19\textwidth, height=0.19\textwidth]{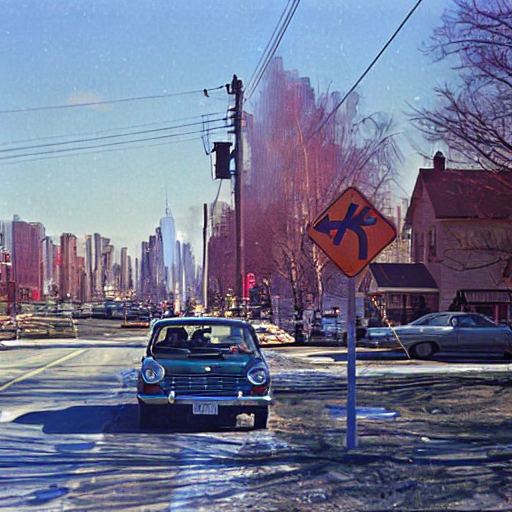} & 
         \includegraphics[width=0.19\textwidth, height=0.19\textwidth]{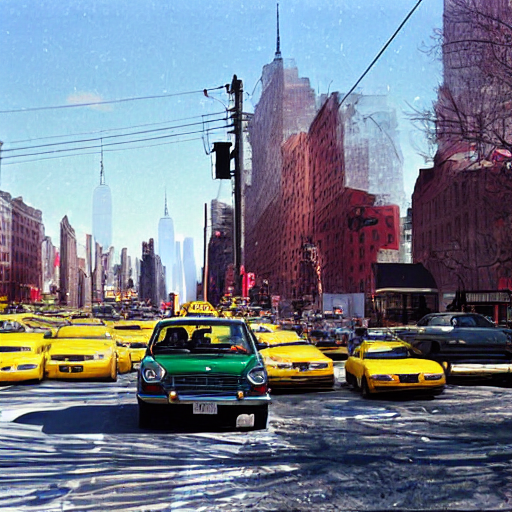} \\
         & Instructions~$\rightarrow$ & \itext{Change this to a New York city street} & \itext{Replace the road with a busy New York city street} & \itext{Add tall buildings in the background} & \itext{Add yellow taxis surrounding the green car}  \\
         & \lines~ \\
         \multicolumn{6}{c}{\vspace{-20pt}}\\
         
         \midrule
         
         \end{tabularx}
    }
    \caption{\textbf{Qualitative results.} Using \methodname on top of IP2P helps to respect the ambiguous instruction (underlined in \textcolor{basecolor}{grey}) by adding specific elements into the input scene. We show how increasing the number of specific instructions (\textcolor{cOne}{orange}, \textcolor{cTwo}{cyan}, \textcolor{cThree}{purple}) adds important details to the scene, ignored by the baseline. Examples of such details are the snow on the ground (first row), the herbs garnish (second row), and the taxis (third row). Colored boxed in the header indicate the instruction used for each column.\looseness=-1}\label{fig:qual}
\end{figure*}

\section{Experiments}\label{sec:experiments}
\subsection{Experimental Setup}\label{sec:setup}
\noindent\textbf{Baselines~~} We test \methodname by adapting the editing diffusion models trained in InstructPix2Pix~\cite{brooks2023instructpix2pix} (IP2P), MB~\cite{zhang2024magicbrush} (MB) and HQEdit~\cite{hui2024hq}. We use the default inference hyperparameters for all methods: $w^\text{C}=7, w^\text{I}=1.5$. We select with a grid search $w^\text{S}=7$ for IP2P, $w^\text{S}=5$ for MB, and $w^\text{S}=9$ for HQEdit. We test with $N=\{1, 2, 3\}$, limiting the number of instructions due to increased computational time associated with higher $N$ values. We incrementally build $\mathcal{S}$ for increasing $N$ values, in such a way that $\mathcal{S}$ with lower $N$ are subsets of those with higher $N$. We use 30 diffusion steps for image generation. Images are $512\times512$.\\ 
\noindent\textbf{Datasets~~}
We test \methodname on two datasets. We first consider the \textit{global} split of the \textit{EMU-Edit} dataset, including 220 real images with ambiguous instructions satisfying our definition~\cite{sheynin2024emu}. We also introduce a subset of 370 images and editing instructions from the IP2P dataset~\cite{brooks2023instructpix2pix}, following related works~\cite{guo2024focus}. We call this set \textit{IP2P data}. To focus our evaluation on ambiguous instructions, we manually classify the 370 samples. We process instructions in both datasets with GPT-4o to extract $\mathcal{S}$.\looseness=-1\\
\noindent\textbf{Metrics~~} We evaluate the quality of edited images with three CLIP-based~\cite{clip} metrics: input image preservation, editing strength and adherence to the edit. First, we use $\text{CLIP}_\text{i}$ to measure input image preservation 
as in~\cite{guo2024focus}, which is the CLIP space cosine similarity between $x$ and $\tilde{x}$. This captures how similar is $\tilde{x}$ to the original $x$. Then, we measure the editing strehgth $\text{CLIP}_d$ as the cosine distance between the CLIP image embedding of $\tilde{x}$ and the text embedding of the final caption, following~\cite{clip}. This assesses the fidelity of the edited image $\tilde{x}$ to the final caption. Finally, we use the directional similarity of~\cite{gal2022stylegan}, referred to as $\text{CLIP}_\Delta$, to measure adherence to the edit. For that, we first process each pair $(x, c)$ with GPT-4o to obtain a short \textit{initial} caption, describing the input scene, and a \textit{final} caption, describing the desired scene after the editing. For instance, if the initial caption is ``\texttt{a woman by the pool}'' and $c$ is ``\texttt{make her a robot}'', a final caption may be ``\texttt{a robot by the pool}''. $\text{CLIP}_\Delta$ is evaluated as the cosine similarity between the difference of the CLIP image embeddings of $x$ and $\tilde{x}$, and the difference of the text embeddings of the initial and final captions. 
This compares the change in the image to the change in the caption.
In addition to these three metrics, we also use TIFA~\cite{hu2023tifa} to evaluate the effectiveness of $s_i \in \mathcal{S}$, as well as LPIPS~\cite{zhang2018unreasonable} and DreamSim~\cite{fu2023dreamsim} to evaluate image diversity. Finally, we evaluate pairwise image preferences with GPT-4o.

\subsection{Editing Performance}\label{sec:exp-editing-performance}
\noindent 
Here, we compare against the state of the art. For each of the three editing diffusion models, we evaluate their performance with and without \methodname applied on top of them. We evaluate models on EMU-Edit and IP2P data both quantitatively (Table~\ref{tab:quant-clip}) and qualitatively (Figure~\ref{fig:qual}).\looseness=-1

\vspace{-7px}
\paragraph{CLIP Metrics} We present results of CLIP-based metrics in Table~\ref{tab:editing-quality}. Overall, we observe performance improvement across all metrics, models and datasets, advocating the advantages of~\methodname. Notably, performance increases with $N$, with models using $N=3$ specific instructions performing best. In particular, we report the biggest improvements in IP2P, where we report for \methodname/Baseline \textbf{0.1858}/0.1203 on EMU-Edit for $N=3$. This suggests that our instruction decomposition helps to follow the ambiguous instruction $c$. Moreover, we observe an increase in image consistency in MB and HQEdit, where we report improvements in $\text{CLIP}_i$ metric. We attribute this behavior to our decomposition strategy. Indeed, while general instructions $c$ alone may lead to ambiguous edits impacting the entire scene, injecting $s_i \in \mathcal{S}$ for inference guides the editing process on spatially-constrained edits. These still convey the desired modifications, as proved by the improvements in $\text{CLIP}_\Delta$ and $\text{CLIP}_\text{d}$. In IP2P, the Baseline reports the highest $\text{CLIP}_\text{i}$. 
While it might seem as if Baseline outperforms all other methods, in reality we observe that IP2P may fail to perform any change to $x$ when the input instruction is too ambiguous, thus artificially inflating the $\text{CLIP}_\text{i}$ metric. This is also confirmed by the lower performance in $\text{CLIP}_d$ and $\text{CLIP}_\Delta$. 

\vspace{-7px}
\paragraph{GPT Evaluation} 
We use GPT-4o to measure pairwise preference for our \methodname against the chosen baselines. 
We choose the configuration with $N=3$, since it yields the best performance in Table~\ref{tab:editing-quality}. For each image-instruction pair $(x, c)$ on each dataset, we prompt GPT to choose the best edited image $\tilde{x}$ between one of the baselines and \methodname. We ask to take into account the fidelity to the editing instruction, the quality and realism of the generated image, and the content preservation from the original picture while making the decision. The original image $x$ is also provided for reference. The prompt is shown in the appendix. We report the average of GPT choices for all model pairs in Table~\ref{tab:gpt-eval}. As before, we significantly outperform all baselines. In particular, we beat the MB baseline on IP2P data (\methodname/Baseline is \textbf{66.0}\%/34.0\%). This is especially interesting, considering that it was the only setup where SANE was not outperforming the baseline $\text{CLIP}_\Delta$ in Table~\ref{tab:editing-quality}, reporting 0.1998/\textbf{0.2028}. 
This result proves that we can benefit from the improved quality of the generated images, even when the fidelity to $c$ is slightly penalized. We attribute this observation to the improved precision of our editing process.\looseness=-1

\vspace{-5px}
\paragraph{Qualitative Results}
We show qualitative results in Figure~\ref{fig:qual}. We use IP2P as a baseline model, and we add \methodname on top of it with $N=\{1, 2, 3\}$ instructions. We use the same hyperparameters and random seed for the baseline and \methodname. \methodname gradually adds details with the specific instructions to respect the ambiguous instruction. Interestingly, adding more specific instructions (higher $N$) brings advantages of removing editing artifacts.
This is especially evident in the third row, where the last two specific instructions help to remove the background artifacts generated by IP2P. We believe such artifacts are generated by the ambiguity of the input instruction. Specific instructions help to constrain the solution space for the editing task, improving the robustness to such undesired visual effects.\looseness=-1

\begin{table}[t]
\begin{subtable}{0.6\linewidth}
\setlength{\tabcolsep}{0.01\linewidth}
\resizebox{\linewidth}{!}{
    \begin{tabular}{cl|c|cc}
    &\multicolumn{1}{c}{}& \multicolumn{1}{c}{}& \multicolumn{2}{c}{\textit{EMU-Edit}} \\
    \toprule
    &\textbf{Method} & \textbf{$N$} & $\text{\textbf{CLIP}}_\text{\textbf{d}}\uparrow$ & $\text{\textbf{CLIP}}_\text{\textbf{$\Delta$}}\uparrow$ \\\midrule
    \multirow{4}{*}{\rotatebox{90}{\textit{IP2P}}} & Baseline & - & 0.2860 & 0.0861 \\\cmidrule{2-5}
    & \multirow{3}{*}{\methodname} & 1 & 0.2854 & 0.1304 \\
    &&  2 & 0.2851 & 0.1436 \\
    &&  3 & \textbf{0.2878} & \textbf{0.1518} \\\midrule
    \multirow{4}{*}{\rotatebox{90}{\textit{MB}}}& Baseline & - & 0.2725 & 0.1108 \\\cmidrule{2-5}
    &\multirow{3}{*}{\methodname} &  1 & 0.2831 & 0.1240 \\
    &&  2 & 0.2876 & 0.1282 \\
    &&  3 & \textbf{0.2926} & \textbf{0.1284} \\\midrule
    \multirow{4}{*}{\rotatebox{90}{\textit{HQEdit}}}& Baseline & - & 0.2522 & 0.1090 \\\cmidrule{2-5}
    &\multirow{3}{*}{\methodname} &  1 & 0.2854 & 0.1238 \\
    &&  2 & 0.2889 & 0.1306 \\
    &&  3 & \textbf{0.2907} & \textbf{0.1326} \\
    \bottomrule
    \end{tabular}
}
\caption{CLIP metrics}\label{tab:interp-clip}
\end{subtable}
\hfill
\begin{subfigure}{0.3\linewidth}
    \includegraphics[width=\linewidth]{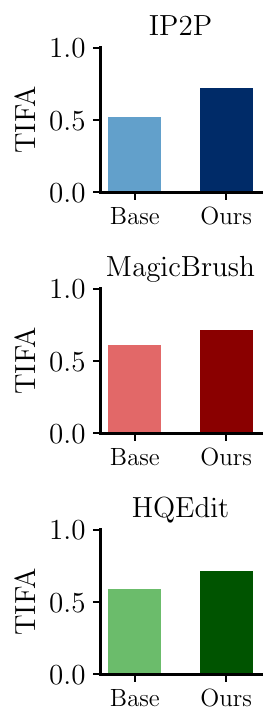}
    \caption{TIFA score}\label{tab:interp-tifa}
\end{subfigure}
\caption{\textbf{Specific instructions effect.} We evaluate how much \methodname and baselines respect the instructions in $\mathcal{S}$. With CLIP metrics~\subref{tab:interp-clip} and TIFA~\subref{tab:interp-tifa}, we verify that we improve fidelity to $\mathcal{S}$.\looseness=-1}\label{tab:interpretability}
\end{table}

\subsection{\methodname Properties}\label{sec:properties}

\begin{figure}[t]
    \centering
    \begin{subfigure}{\linewidth}
    \includegraphics[width=\linewidth]{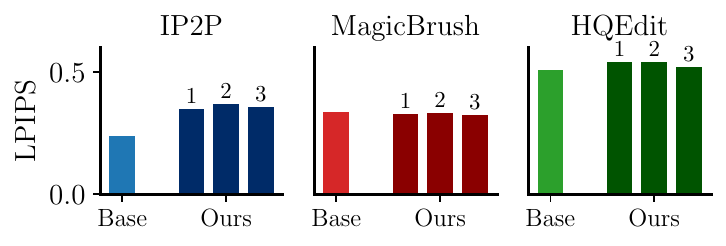}
    \includegraphics[width=\linewidth]{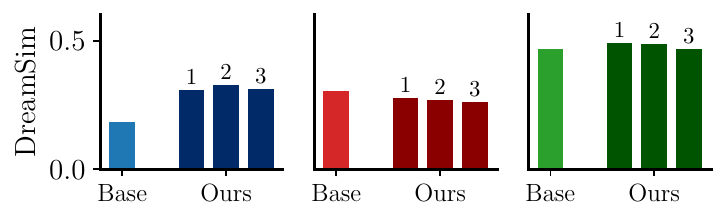}
    \caption{Quantitative evaluation}\label{fig:variability-plots}
    \end{subfigure}
    \begin{subtable}{0.8\linewidth}
    \resizebox{\linewidth}{!}{
    \begin{tabular}{cc|cc}
         & \Large{Image} & \Large{Baseline} & \Large{\methodname}\\
         \raisebox{3.5em}{\rotatebox{90}{\textit{\Large{IP2P}}}} &\includegraphics[width=100px]{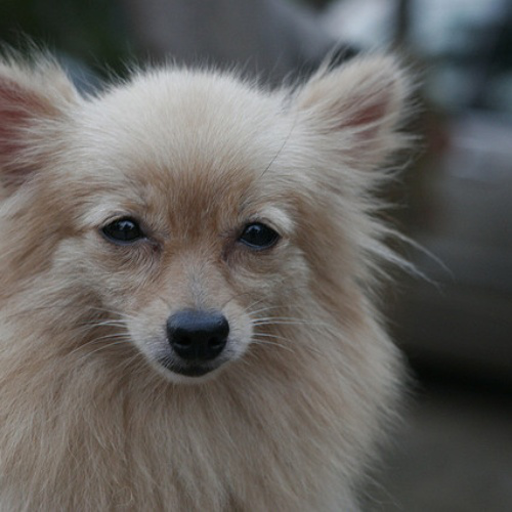} & 
         \includegraphics[width=100px]{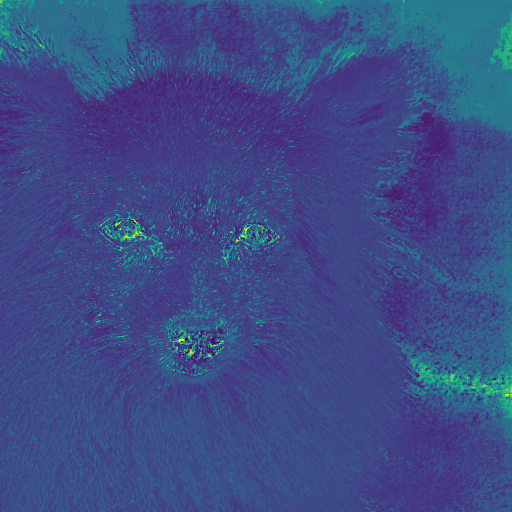} & 
         \includegraphics[width=100px]{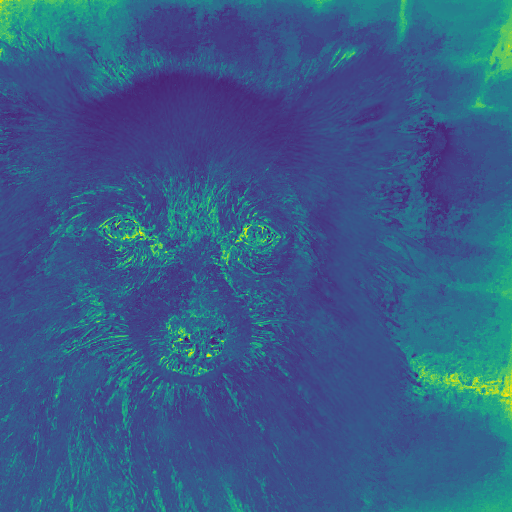} \\
         && \multicolumn{2}{c}{\texttt{``Make the photo seem like it was}}\\
         && \multicolumn{2}{c}{\texttt{taken during a snow storm''}}\\
         
    \end{tabular}
    }
    \caption{Pixel difference visualization}
    \label{fig:variability-viz}
    \end{subtable}
    \caption{\textbf{Variability.} We quantify variability of generated samples with LPIPS and DreamSim, evaluating the average distance to the input image~\subref{fig:variability-plots}. For both, higher values imply higher variability. Above \textit{Ours} bars, we report $N$. In~\subref{fig:variability-viz}, we show the average pixel difference of original and synthesized images for the reported instruction. Using \methodname allows to modify more pixels.}\label{fig:variability}
\end{figure}

\noindent In this section, we discuss additional properties of \methodname, namely  
the impact of instructions in $\mathcal{S}$ (Table~\ref{tab:interpretability}) and its effects on the variability of outputs (Figure~\ref{fig:variability}). We use real images from EMU-Edit for all evaluations.

\paragraph{Effects of Specific Instructions}
To prove that \methodname is working correctly, we need to evaluate whether specific instructions are applied. As mentioned in Section~\ref{sec:method-instruction-decomposition}, 
this would also enable interpretability of the editing process: 
at inference time, we can provide specific instructions $\mathcal{S} = \{s_i\}_{i=1}^{N}$ to the user, to explain how the editing instruction $c$ has been applied. 
For this reason we quantify how much the edited image $\tilde{x}$ respects each instruction in a reference set. We average, for each $\tilde{x}$ and for $N=\{1, 2, 3\}$, $\text{CLIP}_\text{d}$ and $\text{CLIP}_\Delta$ for each $s_i$ in the reference set. We take as reference set the $\mathcal{S}$ obtained with $N=3$, to fairly compare all method and baselines. 
Results in Table~\ref{tab:interp-clip} show that we consistently and considerably outperform all baselines, demonstrating ability of \methodname to effectively apply specific instructions.
As expected, performance increases for higher $N$, with $N=3$ setups consistently outperforming in all scenarios. This means that \methodname is exploiting each instruction in $\mathcal{S}$ for editing. Additionally, we use TIFA~\cite{hu2023tifa} to generate questions related to each $s_i$ with an LLM. We then evaluate if $s_i$ are applied answering the set of question with visual-question answering. For more details, we refer to~\cite{hu2023tifa}. Due to high evaluation costs, we report results only for the setup with $N=3$. Our evaluations in~\ref{tab:interp-tifa} confirm that \methodname correctly exploits the $\mathcal{S}$ instructions.

\vspace{-7px}
\paragraph{Image Variability}%
To evaluate variability of edited images, we randomly select 20 input images from EMU-Edit, and produce 10 edited samples for each of them using \methodname and baselines. We evaluate the mean LPIPS~\cite{zhang2018unreasonable} and DreamSim~\cite{fu2023dreamsim} between input images $x$ and edited images $\tilde{x}$. Higher metrics imply higher variability of the output. LPIPS is particularly sensitive to low-level differences~\cite{zhang2018unreasonable}, while DreamSim captures semantic variability~\cite{fu2023dreamsim}. Results in Figure~\ref{fig:variability-plots} show that we outperform the baseline (\textit{Base}) for IP2P and HQEdit. 
The best performing setups are $N=1$ with HQedit and $N=2$ with IP2P. We speculate that higher $N$ increases the probability of two samples having at least one overlap in generated specific instructions. 
We observe that MB~\cite{zhang2024magicbrush} baseline outputs highly variable outputs, 
which we attribute to the quality of its training set~\cite{zhang2024magicbrush}. We display in Figure~\ref{fig:variability-viz} the average pixelwise difference between $x$ and $\tilde{x}$ for IP2P and IP2P+\methodname. 
\methodname produces edits that are better distributed spatially, including regions ignored by other models (\eg, the dog fur).\looseness=-1

\subsection{Ablation Studies}

\noindent We focus our ablations on instruction combination strategies (Table~\ref{tab:instruction-combination}), also providing insights on our design choices (Table~\ref{tab:arch-components}) 
and on the effectiveness of \methodname on different types of input instructions (Table~\ref{tab:automatic-selection}).\looseness=-1

\vspace{-8px}
\paragraph{Instruction Combination}
We combine specific instructions as presented in Section~\ref{sec:method-instruction-combination}. Here, we study the effectiveness of alternative solutions for instruction combination. We first propose a naive \textit{Prompt Concatenation} baseline, where we concatenate the text of the instruction $c$ with all $s_i \in \mathcal{S}$, using commas as separator. We then perform the editing using the obtained concatenated instruction. We also combine the effects of $c$ and all $s_i \in \mathcal{S}$ with \textit{Composable Diffusion}~\cite{liu2022compositional}. We refer to the original paper~\cite{liu2022compositional} for details. For a fair comparison, we set $w^\text{C}$ and $w^\text{S}$ as weights for $c$ and each $s_i$, respectively. We test with $N\!=\!3$. From results in Table~\ref{tab:instruction-combination}, we infer that our strategy for instruction combination outperforms other strategies. We explain this with the hierarchical nature of instructions: we preserve the fidelity to the original $c$ by design, aggregating the effects of different specific instructions in a single noise estimation.

\begin{table}[t]
    \centering

    \resizebox{0.8\linewidth}{!}{
    \begin{tabular}{cl|ccc}
        & \multicolumn{1}{c}{}& \multicolumn{3}{c}{\textit{EMU-Edit}}\\
    \toprule
     & \textbf{Method} & $\text{\textbf{CLIP}}_\text{\textbf{d}}\uparrow$ & $\text{\textbf{CLIP}}_\text{\textbf{i}}\uparrow$ & $\text{\textbf{CLIP}}_\Delta\uparrow$ \\\midrule
    \multirow{3}{*}{\rotatebox{90}{\textit{IP2P}}} & Prompt concat & 0.2933 & \textbf{0.8188} & 0.1439\\
    & Comp. Diffusion & 0.2944 & 0.7299 & 0.1847 \\
    & Ours & \textbf{0.2968} & 0.7531 & \textbf{0.1858}\\\midrule
    \multirow{3}{*}{\rotatebox{90}{\textit{MB}}} & Prompt concat & 0.2890 & 0.7802 & 0.1639\\
    & Comp. Diffusion & 0.2990 & \textbf{0.8210} & 0.1649 \\
    & Ours & \textbf{0.3006} & 0.8209 & \textbf{0.1655}\\\midrule
    \multirow{3}{*}{\rotatebox{90}{\textit{HQEdit}}} & Prompt concat & 0.2720 & 0.6086 & 0.1512\\
    & Comp. Diffusion & 0.2753 & 0.6547 & 0.1398 \\
    & Ours & \textbf{0.2823} & \textbf{0.6870} & \textbf{0.1474}\\\bottomrule
    \end{tabular}}
\caption{\textbf{Instruction combination baselines.} We outperform two baselines for instruction combination, as shown by CLIP metrics. Only our approach allows to benefit from the distinctions between ambiguous and specific instructions.}\label{tab:instruction-combination}

\end{table}

\vspace{-5px}
\paragraph{Impact of Design Choices}
Here, we analyze the effect of several design choices. First, we study the impact of the initial instruction $c$ on the performance. One may argue that, since each $s_i$ is related to $c$, it may be sufficient to apply $\mathcal{S} = \{s_i\}_{i=1}^{N}$ without $c$ to achieve good editing. We test \methodname on EMU-Edit with $N=3$ and setting $w^T=0$, \ie removing $c$ guidance. Results in Table~\ref{tab:arch-components} (``SANE w/o $c$'') prove that although \methodname w/o $c$ can outperform certain baselines (\eg $\text{CLIP}_\Delta$ = \textbf{0.1552} on IP2P), using $c$ for denoising is important to achieve the best performance. Then, we replace the masking operation described in Equation~\eqref{eq:masking} with a naive averaging of the estimated noises 
$\epsilon_{i,t}^s$. 
In Table~\ref{tab:arch-components}, we show that this alternative strategy, reported as ``\methodname w/ avg'', yields lower performance on CLIP metrics.\looseness=-1

\vspace{-5px}
\paragraph{\methodname with Non-ambiguous Instructions}
\methodname is based on the assumption that the input prompt is ambiguous. We now propose a preliminary evaluation on non-ambiguous instructions to show that: 1) \methodname can also perform well on non-ambiguous prompts; 2) The gain brought by \methodname is higher on ambiguous instructions, thus validating the motivation behind the design of \methodname. 

To this aim, we propose the following experiment. We consider the first set of 1199 images from the IP2P dataset which can be either ambiguous or not. We then use GPT-4o to identify ambiguous instructions leading to 696 ambiguous and 503 specific $(x, c)$ pairs. The prompt is reported in the appendix. For both types of pairs (subsets Amb. and Spec.), we process the samples using IP2P with and without \methodname on top of it, leading to the results in Table~\ref{tab:automatic-selection}. We also report the relative change with respect to the baseline.

Experiment on non-ambiguous instructions show that \methodname achieves  higher adherence to the input edit (higher $\text{CLIP}_\Delta$) while achieving similar preservation of the input image (similar $\text{CLIP}_d$). It demonstrates that \methodname can be effective as a general-purpose editing method, \ie applied to an arbitrary input instruction, and that using LLM can boost the performance.

In addition, the improvement for ambiguous instructions is higher (\eg \textcolor{OliveGreen}{+17.5\%} in $\text{CLIP}_\Delta$) compared to improvement for specific ones (\textcolor{OliveGreen}{+10.8\%}), supporting the motivation for our work. Notably, the absolute performance on specific instructions is higher, confirming our initial observation that editing models have difficulties with ambiguous inputs. \looseness=-1

\begin{table}[t]
\setlength{\tabcolsep}{0.02\linewidth}
\centering
\resizebox{0.85\linewidth}{!}{
    \begin{tabular}{c@{\hspace{0.05\linewidth}}lH@{\hspace{0.1\linewidth}}|ccc}
    \multicolumn{3}{c}{}& \multicolumn{3}{c}{\textit{EMU-Edit}}\\
    \toprule
    & \textbf{Method} & \textbf{$N$} & $\text{\textbf{CLIP}}_\text{\textbf{d}}\uparrow$ & $\text{\textbf{CLIP}}_\text{\textbf{i}}\uparrow$ & $\text{\textbf{CLIP}}_\Delta\uparrow$ \\\midrule
    \raisebox{0px}{\multirow{4}{*}{\rotatebox{90}{\textit{IP2P}}}}& Baseline & - & 0.2923 & \textbf{0.8810} & 0.1203\\
    \multirow{1}{*}{} & \multirow{1}{*}{\methodname w/o $c$} &3& 0.2883 & 0.7796 & 0.1552 \\
    \multirow{1}{*}{} & \multirow{1}{*}{\methodname w/ $\text{avg}$} &3& \textbf{0.2974} & \underline{0.8190} & \underline{0.1587} \\
    &\methodname & & \underline{0.2968}& 0.7531 & \textbf{0.1858}\\
    \midrule
    \raisebox{0px}{\multirow{4}{*}{\rotatebox{90}{\textit{MB}}}}& Baseline & - & 0.2888 & 0.7858 & 0.1618 \\
    \multirow{1}{*}{} & \multirow{1}{*}{\methodname w/o $c$} &3& 0.2954 & \textbf{0.8795} & 0.1375 \\
    \multirow{1}{*}{} & \multirow{1}{*}{\methodname w/ $\text{avg}$} &3& \underline{0.2978} & 0.8087 & \textbf{0.1681} \\
    &\methodname&&\textbf{0.3006} &\underline{0.8209} &\underline{0.1655}\\
    \midrule
    \raisebox{0px}{\multirow{4}{*}{\rotatebox{90}{\textit{HQEdit}}}}& Baseline & - & 0.2675 & 0.6501 & 0.1417\\
    \multirow{1}{*}{} & \multirow{1}{*}{\methodname w/o $c$} &3&0.2795 & \textbf{0.6983} & 0.1321 \\
    \multirow{1}{*}{} & \multirow{1}{*}{\methodname w/ $\text{avg}$} &3&\underline{0.2807} & 0.6852 & \underline{0.1442} \\
    &\methodname&&\textbf{0.2823}& \underline{0.6870} & \textbf{0.1474}\\
    \bottomrule
    \end{tabular}
}
\caption{\textbf{Design choices.} Removing the ambiguous instruction $c$  (SAGE w/o $c$) results in degraded performance, proving that editing models benefit from both specific and ambiguous instructions. Also, replacing the masking operation with averaging (SAGE w/ avg) results in worse editing. We highlight \textbf{first} and \underline{second} best.\looseness=-1}\label{tab:arch-components}
\end{table}

\begin{table}[t]
\setlength{\tabcolsep}{0.02\linewidth}
\centering
\resizebox{\linewidth}{!}{
    \begin{tabular}{c@{\hspace{0.05\linewidth}}cl|lll}
    \toprule
    & \textbf{Prompts} & \textbf{Method} & \multicolumn{1}{c}{$\text{\textbf{CLIP}}_\text{\textbf{d}}\uparrow$} & \multicolumn{1}{c}{$\text{\textbf{CLIP}}_\text{\textbf{i}}\uparrow$} & \multicolumn{1}{c}{$\text{\textbf{CLIP}}_\Delta\uparrow$} \\\midrule
    \raisebox{-3px}{\multirow{4}{*}{\rotatebox{90}{\textit{IP2P}}}}&\multirow{2}{*}{Amb.} & Baseline & 0.2946 & \textbf{0.8992} & 0.1504\\
    && \methodname & \textbf{0.2998}~\diff{+1.76\%} & 0.8225~\diff{-8.53\%} & \textbf{0.1767}~\diff{+17.5\%} \\\cmidrule{2-6}
    &\multirow{2}{*}{Spec.} & Baseline & \textbf{0.3044} & \textbf{0.9000} & 0.2012 \\
    && \methodname & \textbf{0.3044}~\small{(+0.00\%)} & 0.8302~\diff{-7.75\%} & \textbf{0.2228}~\diff{+10.8\%}\\
    \bottomrule
    \end{tabular}
}
\caption{\textbf{Evaluation on non-ambiguous prompts.} We report beneficial effects of \methodname on both ambiguous (Amb.) and specific instructions (Spec.)\looseness=-1}\label{tab:automatic-selection}

\end{table}

\section{Conclusions}\label{sec:conclusions}

We have introduced \textit{Specify ANd Edit} (\methodname), a zero-shot inference pipeline that improves the performance of diffusion-based text-to-image editing methods with ambiguous instructions. By using an LLM to decompose instructions into specific interventions, \methodname enhances both interpretability and editing quality. Our experiments show consistent performance improvements and increased output diversity. Moreover, \methodname is versatile, and it can benefit both ambiguous and clear editing tasks. In the future, we plan to address the limitations of \methodname, such as the difficulty in handling a high number of specific instructions and the lack of guarantee that each specific instruction is actually applied (see appendix).\looseness=-1

\section*{Acknowledgements}
EI and SL are supported by the French National Research Agency (ANR) with the ANR-20-CE23-0027 project. FP is funded by KAUST (Grant DFR07910). PT is supported by UKRI grant: Turing AI Fellowship EP/W002981/1, and by the Royal Academy of Engineering under the Research Chair and Senior Research Fellowships scheme.

{\small
\bibliographystyle{ieee_fullname}
\bibliography{main}
}

\clearpage
\appendix
\input{sections_supp/supp_intro}
\section{Limitations}\label{sec:supp-limitations}
In this section, we discuss the limitations of \methodname. First, despite the observation that specific instructions $\mathcal{S}=\{s_{i}\}_{i=1}^{N}$ improve model performance on the initial instruction $c$, there is no guarantee that all of these specific instructions will always be respected. In our experience, the hire is $N$, the more the editing process tends to avoid following one or more specific instructions $s_{i}$. This is expected, since increasing the number of instructions makes the editing task more challenging for the diffusion model. To provide a transparent evaluation, we include an additional experiment quantifying how a single editing instruction is respected in Section~\ref{sec:supp-exp}. Moreover, \methodname brings additional computational costs due to multiple denoising operations. We quantify these effects as well in Section~\ref{sec:supp-exp}.

\section{Prompts}\label{sec:supp-prompts}
Here, we report all the prompts used in the paper. To make it easier to understand our prompt engineering choices, we color differently the parts of the prompts designed for encouraging specific behaviour in the LLM. In particular, we use \textcolor{KleinBlue}{blue} for indicating the task-related instruction to the model, \textcolor{orange}{orange} for parsing-related instructions, \textcolor{OliveGreen}{green} for in-context learning, and \textcolor{DarkRed}{red} for additional task-specific instructions to mitigate the presence of errors, established via trial-and-error.

\paragraph{Instruction Decomposition}
Our instruction decomposition prompt used in Section~\ref{sec:method-instruction-decomposition} is reported below. The \texttt{<caption>} is obtained by using the captioning prompt, shown in the next paragraph.

\begin{tcolorbox}[colback=gray!5!white, colframe=gray!75!black, title=\textbf{Decomposition prompt}, breakable, boxrule=0.5mm, colbacktitle=gray!75!black, coltitle=white]

\textcolor{KleinBlue}{
You are a helpful assistant for image editing. 
I will provide you with a caption that describes an image,
and an editing instruction that represents an ambiguous modification
of the scene. Your task is to propose specific modifications. 
You can ask to add or replace elements in the scene,
proposing consistent modification that agree with the
ambiguous instruction.}\\

\textcolor{orange}{Be concise and output your instructions without
further considerations or reasoning, one local modification per line.
Do not output any other text than the suggested outputs, do not write
``suggested output:''.\looseness=-1}\\

\textcolor{OliveGreen}{I am going to provide some examples now.\\
Caption: a photo of a urban scenario, with cars.\\ 
ambiguous instruction: make the scene vintage.\\
Suggested output: replace the cars with old cars\\
Caption: a photo of a dog running on the grass.\\ 
ambiguous instruction: make it look funny.\\ 
Suggested output: add a hat to the dog.}\\

\textcolor{DarkRed}{The main subject of the scene must stay the same. For instance, if the photo is describing a cat
as the main subject, you cannot replace the cat with} \textcolor{DarkRed}{another animal. You should NEVER remove elements.
Only propose instructions targeting elements that appear in the caption, without imagining anything else.\looseness=-1}\\

\textcolor{KleinBlue}{Now, provide \texttt{<$N$>} outputs for the following caption and subjective instruction.
Caption: \texttt{<caption>} ambiguous instruction 
\texttt{<$c$>}}

\end{tcolorbox}

\paragraph{Captioning prompt}
To evaluate $\text{CLIP}_\Delta$ (Section~\ref{sec:setup}), we generate simultaneously the captions of the initial and the final scenes with GPT-4o. Please note that we use only the caption of the original image in our inference pipeline (Section~\ref{sec:method}). 
However, we observe that generating both captions simultaneously improves their consistency, allowing us to better evaluate $\text{CLIP}_\Delta$ in Section~\ref{sec:experiments}. 
Here, we report the prompt used for the captioning task. We input the ambiguous instruction $c$, along with the input image $x$. The prompt is:

\begin{tcolorbox}[colback=gray!5!white, colframe=gray!75!black, title=\textbf{Captioning prompt}, breakable, boxrule=0.5mm, colbacktitle=gray!75!black, coltitle=white]

\textcolor{KleinBlue}{I am going to provide an input image and an editing instruction. You should propose 1) a caption that describes accurately the input image, max 10 words, focusing only on visual content 2) a caption that encompasses how the image should look like after applying the instruction. The instruction is: \texttt{<$c$>}. The image is \texttt{<$x$>.}}\\

\textcolor{DarkRed}{Try to keep these captions as compact as possible. The captions should be as similar as possible to each other.}\\

\textcolor{orange}{
You should reply following the format: \\1. ``caption 1''\\ 2. ``caption 2'' \\
Just reply with the captions without reasoning or considerations. }
\end{tcolorbox}

\paragraph{GPT evaluation}
Here, we report the prompt used for evaluating image quality with GPT-4o (Section~\ref{sec:exp-editing-performance}).
We indicate with $\tilde{x}_\text{baseline}$ and $\tilde{x}_\text{ours}$ the editing results of baselines and \methodname, respectively.

\begin{tcolorbox}[colback=gray!5!white, colframe=gray!75!black, title=\textbf{GPT evaluation prompt}, breakable, boxrule=0.5mm, colbacktitle=gray!75!black, coltitle=white]

\textcolor{KleinBlue}{I'm going to provide three pictures and one editing textual instruction. The first is an original picture.
The second and the third pictures are edited pictures, where image editing methods are applying transformations to the original picture by following the instruction.
The image editing methods identifiers are A and B. You should tell me what is the editing method that produces the best edited image.}
\textcolor{KleinBlue}{For your evaluation, you should balance how much the edited image respects the instruction, the quality and realism of the generated image, and the content preservation from the original picture.}\\

\textcolor{orange}{Reply with A or B only without further text.
The images are ordered in this way: original image, the image of method A, the image of method B.}\\

\textcolor{KleinBlue}{Now, provide your answer for the input images and the instruction: \texttt{<$c$>} images: \texttt{<$x$>}, \texttt{<$\tilde{x}_\text{baseline}$>, \texttt{<$\tilde{x}_\text{ours}$>}}.}

 \end{tcolorbox}

 \paragraph{Ambiguous instruction selection}
Here, we present the prompt used to automatically select samples with abstract instructions for Table~\ref{tab:automatic-selection}.

\begin{tcolorbox}[colback=gray!5!white, colframe=gray!75!black, title=\textbf{Instruction selection prompt}, breakable, boxrule=0.5mm, colbacktitle=gray!75!black, coltitle=white]
\textcolor{KleinBlue}{You are a helpful assistant for image editing.
I will provide you with an editing instruction that requires certain modification of the scene in the image.
Your task is to decide whether this instruction represents an abstract instruction or a specific instruction.}\\

\textcolor{OliveGreen}{
Here are some examples:\\
ambiguous instruction: ‘Change the image so it apears old and musty.’\\
ambiguous instruction: ‘Make it a snowy day.’\\
ambiguous instruction: ‘Change the image so the players look like zombies.’\\
Specific instruction: ’Add the word ‘tray’ in white to the bottom of the image.'\\
Specific instruction: ‘Change the sheep into a calf.’\\
Specific instruction: ‘Draw this in an oil painting style.’\\}

\textcolor{KleinBlue}{Now, tell me if the following instruction is ambiguous or specific. The instruction is: \texttt{<$c$>}.}\\

\textcolor{DarkRed}{Before answering, motivate your decision by reasoning about the properties of this instruction.}\\

\textcolor{orange}{Your response should start with either ‘Response: ambiguous.’ or ‘Response: specific.}
\end{tcolorbox}

\section{Additional Results}\label{sec:supp-exp}
In this section we provide complementary results and analysis for various aspects of the proposed pipeline.
First, we provide additional evaluation of the model fidelity to specific instructions $\mathcal{S}$. Then, we share insights on the embedding distribution for ambiguous and specific instructions. Finally, we provide additional qualitative results for all methods discussed in Section~\ref{sec:experiments}.\looseness=-1

\begin{table}[t]
\centering    

\setlength{\tabcolsep}{0.01\linewidth}
\resizebox{0.9\linewidth}{!}{
    \begin{tabular}{c@{\hspace{0.05\linewidth}}lc@{\hspace{0.1\linewidth}}|ccc}
    \multicolumn{3}{c}{}& \multicolumn{3}{c}{\textit{EMU-Edit}}\\
    \toprule
    & \textbf{Method} & \textbf{LLM} & $\text{\textbf{CLIP}}_\text{\textbf{d}}\uparrow$ & $\text{\textbf{CLIP}}_\text{\textbf{i}}\uparrow$ & $\text{\textbf{CLIP}}_\Delta\uparrow$ \\\midrule
    \raisebox{-3px}{\multirow{4}{*}{\rotatebox{90}{\textit{IP2P}}}}& Baseline & - & 0.2923 & \textbf{0.8810} & 0.1203\\\cmidrule{2-6}
    \multirow{3}{*}{} & \multirow{3}{*}{\methodname} &GPT-4o& \textbf{0.2968} & 0.7531 & \textbf{0.1858} \\
    &  &LLaMA3-instruct & 0.2903 & 0.7597& 0.1657\\
    & & Mistral v0.3 & 0.2955 & 0.7520 & 0.1771\\
    \midrule
    
    \raisebox{-0px}{\multirow{4}{*}{\rotatebox{90}{\textit{MB}}}}& Baseline & - & 0.2888 & 0.7858 & 0.1618\\\cmidrule{2-6}
    \multirow{3}{*}{} & \multirow{3}{*}{\methodname} &GPT-4o& \textbf{0.3006} & 0.8209 & \textbf{0.1655}  \\
    &  &LLaMA3-instruct & 0.3001 & \textbf{0.8299} & 0.1605\\
    & & Mistral v0.3 & 0.2943 & 0.7977 & 0.1688\\
    \midrule
    
    \raisebox{-3px}{\multirow{4}{*}{\rotatebox{90}{\textit{HQEdit}}}}& Baseline & - & 0.2675 & 0.6501 & 0.1417 \\\cmidrule{2-6}
    \multirow{3}{*}{} & \multirow{3}{*}{\methodname} &GPT-4o& \textbf{0.2823} & \textbf{0.6870} & \textbf{0.1474} \\
    &  &LLaMA3-instruct &0.2808 & 0.6793 & 0.1336\\
    & & Mistral v0.3 & 0.2772 & 0.6521 & 0.1394\\
    \midrule

    \end{tabular}
}
\caption{\textbf{Performance with other LLMs.} Open source alternatives to GPT-4o such as LLaMA3-instruct and Mistral v0.3 perform competitively on the instruction decomposition task.}\label{tab:open-source-models}

\end{table}

\paragraph{Impact of the Language Model}
We replace GPT-4o with LLaMA3-instruct~\cite{meta2024llama3} and Mistral v0.3~\cite{mistral2024mistral7b}, and evaluate \methodname with $N=3$. 
Results in Table~\ref{tab:open-source-models} report only slight decrease in performance. 
Interestingly, decomposing instructions wtih LLaMA3-instruct seems to promote image consistency, outperforming GPT-4o on $\text{CLIP}_{i}$ using IP2P (\textbf{0.7597}) and MB (\textbf{0.8299}). More important, this proves that \methodname can be used in conjunction with open source models, 
promoting accessibility and reproducibility of our results.

\paragraph{How much each specific instruction is respected?}
In Section~\ref{sec:properties} of the main paper, we highlight that the fidelity to specific instructions is an important property of \methodname. Indeed, if we respect specific instructions during editing, we can provide them to the user, improving interpretability of the editing process. In the main paper (Table~\ref{tab:interpretability}), we consider the reference sets $\mathcal{S}$  of specific instructions with $N=3$, and compare the average fidelity of the baselines and the proposed \methodname with $N=\{1, 2, 3\}$ to each of those pre-defined $3$ instructions. 
This allows us to conclude that each specific instruction has an impact on the final image transformation represented by that reference set $\mathcal{S}$ with $N=3$. 
However, we also need to evaluate for each $N$ the fidelity of \methodname to the set of specific instructions actually being used in the denoising process.
In a similar vein to Table~\ref{tab:interpretability}, in Table~\ref{tab:interpretability-single} we report the average over $N$ CLIP-based metrics with respect to each specific instruction $s_{i}$, but considering only the corresponding set of $N$ instructions. 
This means that we evaluate fidelity with respect to a single specific instruction if $N=1$, to two if $N=2$, and to three if $N=3$. Considering that the baselines do not use specific instructions for inference, it is impossible to compare with them in this setup, thus motivating the setup chosen for Table~\ref{tab:interpretability}. As visible from the results, the fidelity to each specific instruction decreases with higher $N$. This is expected, since with more instructions to follow, the editing task is more challenging, and the editing effects might overlap. 
However, we highlight that the $N=3$ setup still performs best in Table~\ref{tab:quant-clip}, resulting in a trade-off for the choice of $N$. We empirically find $N=3$ to be a good value.

\begin{table}[t]
\setlength{\tabcolsep}{0.05\linewidth}
\resizebox{\linewidth}{!}{
    \begin{tabular}{cl|c|cc}
    &\multicolumn{1}{c}{}& \multicolumn{1}{c}{}& \multicolumn{2}{c}{\textit{EMU-Edit}} \\
    \toprule
    &\textbf{Method} & \textbf{$N$} & $\text{\textbf{CLIP}}_\text{\textbf{d}}\uparrow$ & $\text{\textbf{CLIP}}_\text{\textbf{$\Delta$}}\uparrow$ \\\midrule
    \multirow{3}{*}{\rotatebox{90}{\textit{IP2P}}} & \multirow{3}{*}{\methodname} & 1 & \textbf{0.2990} & \textbf{0.1679}  \\
    &&  2 & 0.2907 & 0.156 \\
    &&  3 & 0.2878 & 0.1518 \\\midrule
    \multirow{3}{*}{\rotatebox{90}{\textit{MB}}} & \multirow{3}{*}{\methodname} &  1 & \textbf{0.2930} & \textbf{0.1551} \\
    &&  2 & 0.2924 & 0.1408 \\
    &&  3 &  0.2926 & 0.1284 \\\midrule
    \multirow{3}{*}{\rotatebox{90}{\textit{HQEdit}}}& \multirow{3}{*}{\methodname} &  1 & \textbf{0.2970} & \textbf{0.1593} \\
    &&  2 & 0.2944 & 0.1455 \\
    &&  3 & 0.2907 & 0.1326 \\
    \bottomrule
    \end{tabular}
}
\caption{\textbf{Impact of all the applied specific instructions} We evaluate how much the applied specific instructions influence the performance of \methodname. As expected, with an increasing number of instructions, the fidelity to all the instruction set $\mathcal{S}$ decreases.}\label{tab:interpretability-single}
\end{table}

\begin{table}[t]
\setlength{\tabcolsep}{0.01\linewidth}
\resizebox{\linewidth}{!}{
    \begin{tabular}{cl|ccc|ccc}
    &\multicolumn{1}{c}{}& \multicolumn{3}{c}{\textit{IP2P data}$\rightarrow$\textit{EMU-Edit}} & \multicolumn{3}{c}{\textit{EMU-Edit}$\rightarrow$\textit{IP2P data}}\\
    \toprule
    &\textbf{Method} & $\text{\textbf{CLIP}}_\text{\textbf{d}}\uparrow$ & $\text{\textbf{CLIP}}_\text{\textbf{i}}\uparrow$ & $\text{\textbf{CLIP}}_\Delta\uparrow$ & $\text{\textbf{CLIP}}_\text{\textbf{d}}\uparrow$ & $\text{\textbf{CLIP}}_\text{\textbf{i}}\uparrow$ & $\text{\textbf{CLIP}}_\Delta\uparrow$\\\midrule    \multirow{2}{*}{\rotatebox{90}{\textit{IP2P}}} & Baseline & 0.2906 & \textbf{0.8789} & 0.1076 & 0.2937 & \textbf{0.9084} & 0.1745 \\
    & \methodname & \textbf{0.2939} & 0.7828 & \textbf{0.1468} & \textbf{0.2985} & 0.837 & \textbf{0.2005}\\\midrule
    \multirow{2}{*}{\rotatebox{90}{\textit{MB}}}& Baseline & 0.292 & 0.8163 & 0.142 & 0.2872 & 0.8067 & \textbf{0.186} \\
    & \methodname & \textbf{0.2998} & \textbf{0.8301} & \textbf{0.1457} & \textbf{0.296} & \textbf{0.8297} & 0.1816\\\midrule
    \multirow{2}{*}{\rotatebox{90}{\footnotesize{\textit{HQEdit}}}} & Baseline & 0.2662 & 0.6765 & 0.1184 & 0.2747 & 0.7226 & \textbf{0.1754} \\
    & \methodname & \textbf{0.2853} & \textbf{0.6904} & \textbf{0.1288} & \textbf{0.2895} & \textbf{0.7435} & 0.1735\\
    \bottomrule
    \end{tabular}
}
\caption{\textbf{Evaluation on cross-datasets.} For each dataset, we select a new cross-dataset of ambiguous instructions using the nearest neighbor classifier fitted on the other dataset. The results demonstrate the robustness of \methodname across different datasets}\label{tab:cross_datasets_results}
\end{table}

\begin{figure}[t]
    \centering
    \includegraphics[width=\linewidth]{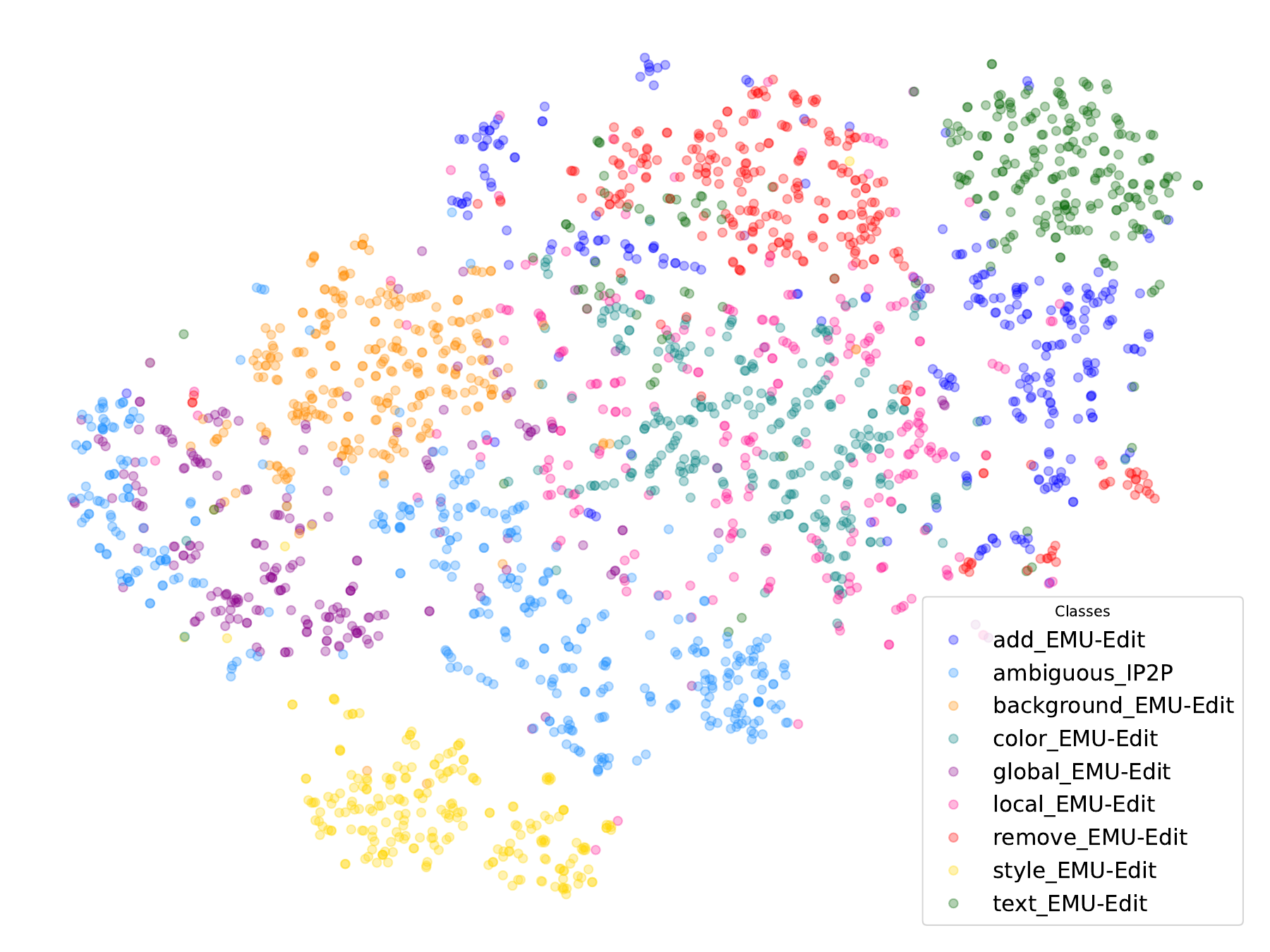}
    \caption{\textbf{t-SNE visualisation of instruction embeddings.} We compute embeddings for all instructions from the EMU-Edit dataset and for ambiguous instructions from the IP2P dataset, and we visualise them using t-SNE. We show that embeddings form clusters corresponding to the task types (splits) in EMU-Edit.}
    \label{fig:tsne_ambiguous}
\end{figure}

\paragraph{Analysis of instruction embeddings}
In Section~\ref{sec:experiments} we extensively evaluate our \methodname on two datasets: EMU-Edit~\cite{sheynin2024emu} and IP2P data~\cite{brooks2023instructpix2pix}. To analyze the instruction space of these datasets, we embed all instructions into the vectors of length $3072$ using \texttt{text-embedding-3-large} by OpenAI, and visualise these embeddings using t-SNE~\cite{van2008visualizing}. In Figure~\ref{fig:tsne_ambiguous}, we show all splits from EMU-Edit against ambiguous instructions from IP2P data. Interestingly, instruction embeddings from EMU-Edit form distinct clusters corresponding to the types of the task (splits). This shows that the instruction space has a complex structure which depends on the semantics encoded by instructions. Moreover, the set of ambiguous instructions from EMU-Edit (\textit{global\_EMU-Edit}) is completely included in the set of ambiguous instructions from IP2P (\textit{ambiguous\_IP2P}), showing the affinity of these sets. To evaluate cross-dataset robustness of our \methodname, we take advantage of this affinity, and perform cross-dataset ambiguous instruction classification by using ambiguous instructions from one dataset to fit a nearest neighbor classifier, and by selecting with this classifier the ambiguous instructions from the other dataset. We denote the selected instructions as ``cross-datasets'' \textit{IP2P data}$\rightarrow$\textit{EMU-Edit} (for EMU-Edit classified with IP2P data) and \textit{EMU-Edit}$\rightarrow$\textit{IP2P data} (for IP2P classified with EMU-Edit). We evaluate our \methodname on these cross-datasets for $N=3$ and compare it against the same baselines as in Section~\ref{sec:experiments}. The results in Table~\ref{tab:cross_datasets_results} not only demonstrate the robustness of our model across different datasets, but also show that such cross-dataset classification can be used as an alternative ambiguous instruction selection strategy that does not require any prompting.
 
\begin{figure}[t]
    \centering
    \includegraphics[width=\linewidth]{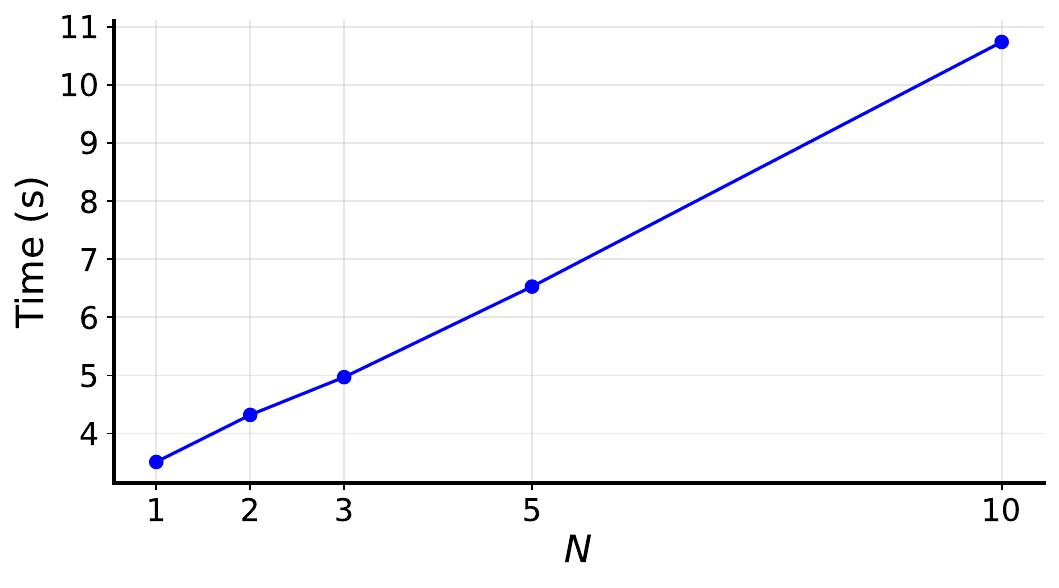}
    \caption{\textbf{Computational costs for different $N$.} We measure the average time for editing one image with multiple $N$ configurations. $N$ and processing times are directly proportional.}
    \label{fig:compute}
\end{figure}

\paragraph{Computational times}
\methodname implies an increased computational load due to multiple denoising operations required for processing specific instructions. We measure the time required to produce an edited image $\tilde{x}$ depending on the number of specific instructions $N$. For each point in Figure~\ref{fig:compute}, we average the processing time over 10 images, using InstructPix2Pix as baseline on NVIDIA 4090 GPU. As seen, increasing $N$ brings a considerable computation overload. This, in addition to results in Table~\ref{tab:interpretability-single}, 
justify the schoice to limit the number of specifc instructions to $N=\{1, 2, 3\}$ in our work.

\begin{figure*}[t]
    \centering
    \setlength{\tabcolsep}{0pt} %
    \resizebox{\linewidth}{!}{
    \begin{tabularx}{\textwidth}{@{}H@{\hspace{0.01\textwidth}}Y@{\hspace{0.01\textwidth}}Y@{\hspace{0.01\textwidth}}YYY@{}}
    & & & \multicolumn{3}{c}{\methodname}\\
    & & Baseline & \small{$N=1$} & \small{$N=2$} & \small{$N=3$}\\
    \cmidrule(lr){3-3}\cmidrule(lr){4-6}
         & Original image & \sqHeader~ \\\midrule
         \raisebox{-5pt}{\rotatebox{90}{\textit{IP2P}}}
         & \includegraphics[width=0.19\textwidth, height=0.19\textwidth]{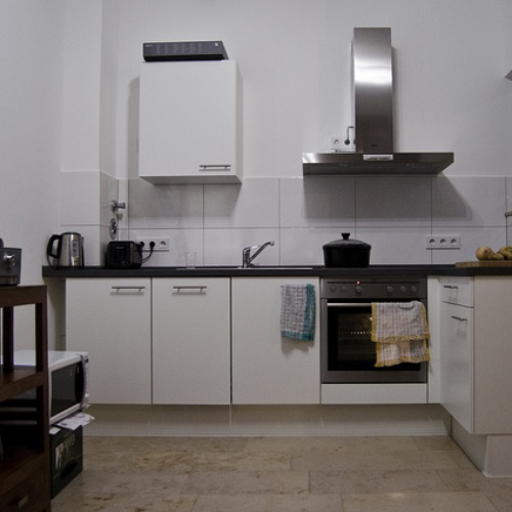} & 
         \includegraphics[width=0.19\textwidth, height=0.19\textwidth]{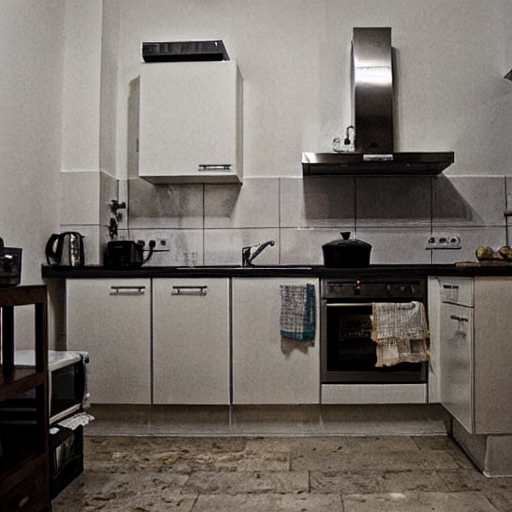} & 
         \includegraphics[width=0.19\textwidth, height=0.19\textwidth]{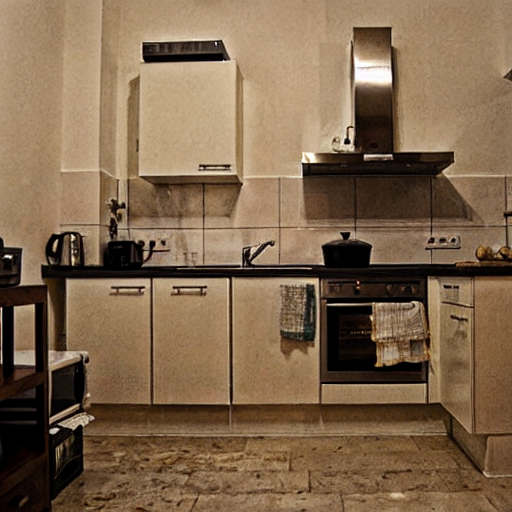} & 
         \includegraphics[width=0.19\textwidth, height=0.19\textwidth]{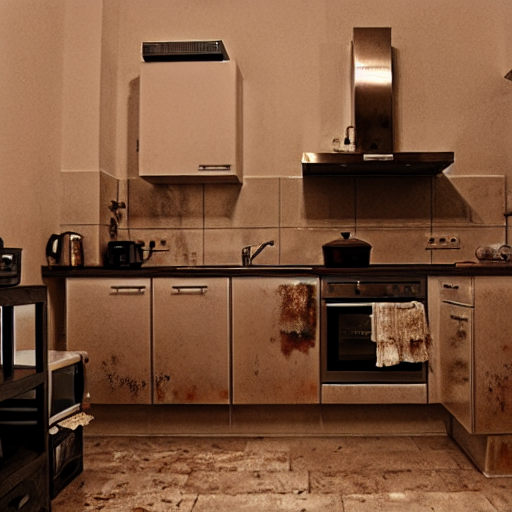} & 
         \includegraphics[width=0.19\textwidth, height=0.19\textwidth]{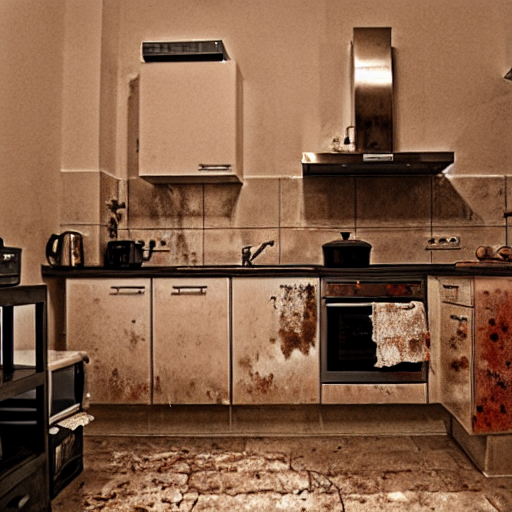} \\
         & Instructions~$\rightarrow$ & \itext{Change the image so it apears old and musty.} & \itext{Apply a sepia filter to the entire image} & \itext{Add visible rust to the refrigerator} & \itext{Add cracks and stains to the stove}  \\
         & \lines~ \\
         \multicolumn{6}{c}{\vspace{-20pt}}\\
         
         \midrule
         \raisebox{-20pt}{\rotatebox{90}{\textit{MagicBrush}}}
         & \includegraphics[width=0.19\textwidth, height=0.19\textwidth]{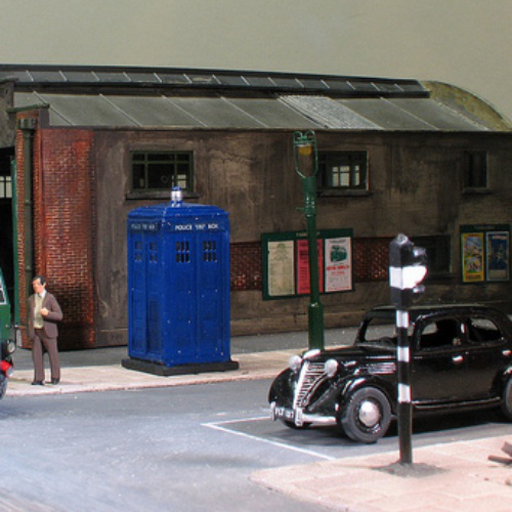} & 
         \includegraphics[width=0.19\textwidth, height=0.19\textwidth]{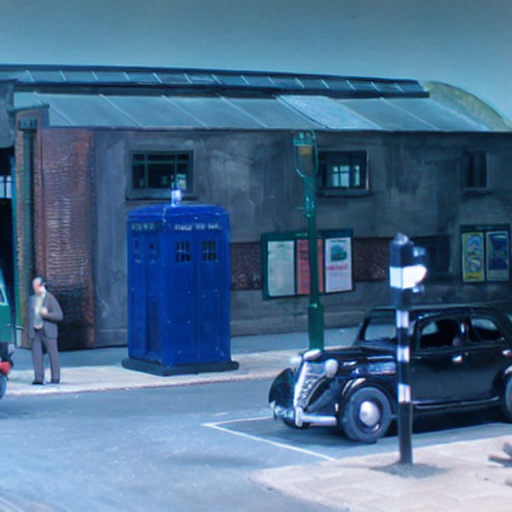} & 
         \includegraphics[width=0.19\textwidth, height=0.19\textwidth]{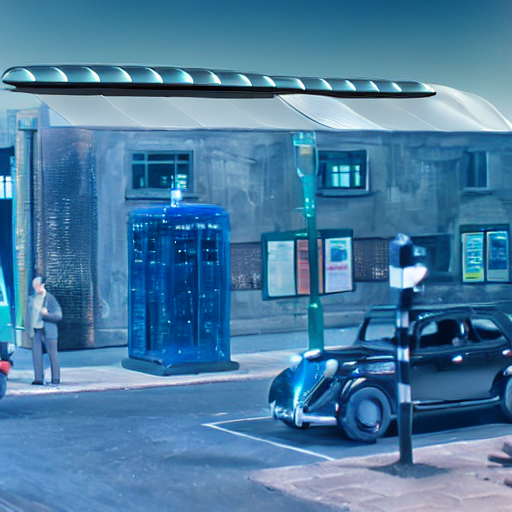} & 
         \includegraphics[width=0.19\textwidth, height=0.19\textwidth]{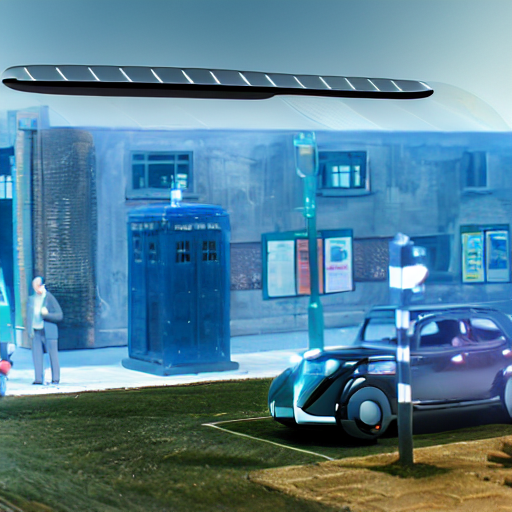} & 
         \includegraphics[width=0.19\textwidth, height=0.19\textwidth]{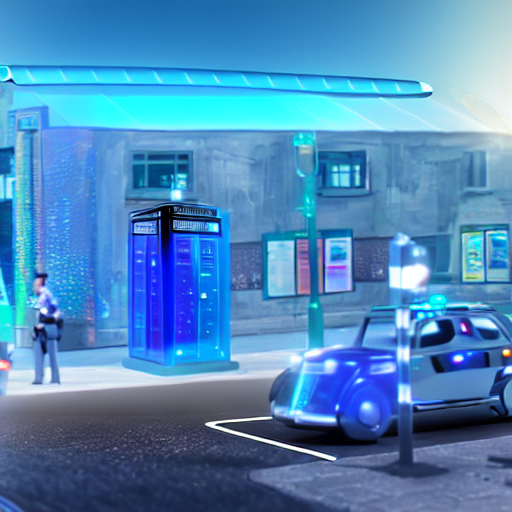} \\
         & Instructions~$\rightarrow$ & \itext{Change the image so it appears to be set in the distant future} & \itext{Add futuristic buildings in the background} & \itext{Replace the road with a modern hover road} & \itext{Add holographic signs around the police box}  \\
         & \lines~ \\
         \multicolumn{6}{c}{\vspace{-20pt}}\\
         
         \midrule

         \raisebox{-10pt}{\rotatebox{90}{\textit{HQEdit}}}
         & \includegraphics[width=0.19\textwidth, height=0.19\textwidth]{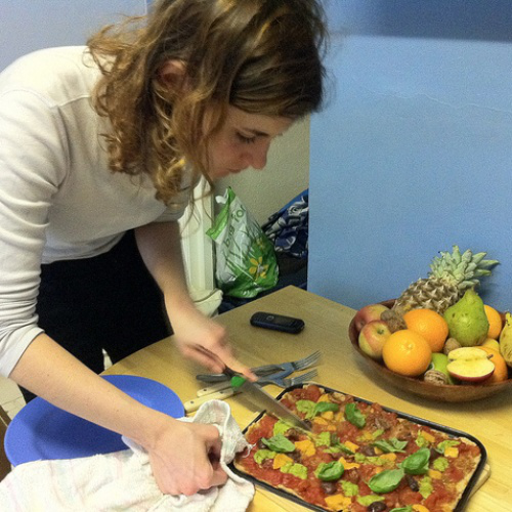} & 
         \includegraphics[width=0.19\textwidth, height=0.19\textwidth]{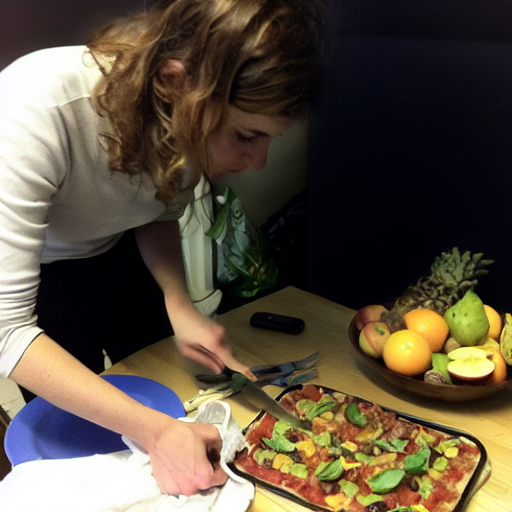} & 
         \includegraphics[width=0.19\textwidth, height=0.19\textwidth]{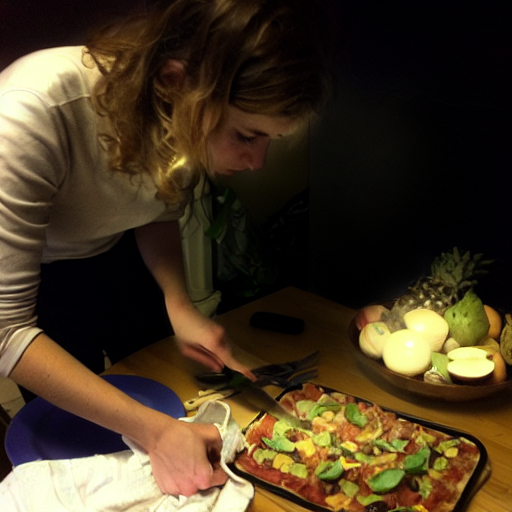} & 
         \includegraphics[width=0.19\textwidth, height=0.19\textwidth]{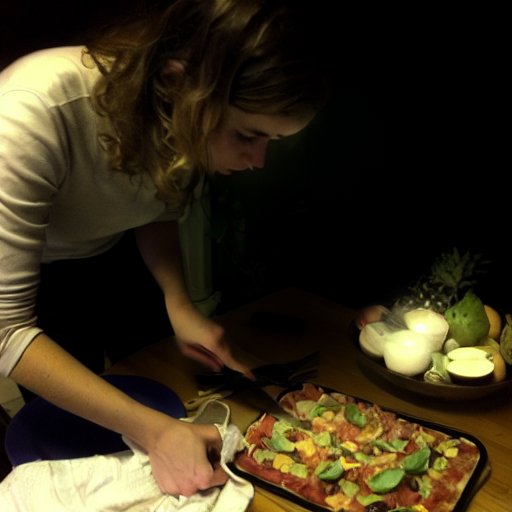} & 
         \includegraphics[width=0.19\textwidth, height=0.19\textwidth]{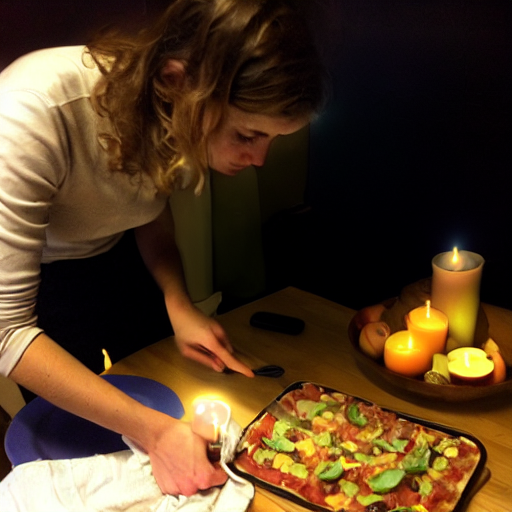} \\
         & Instructions~$\rightarrow$ & \itext{Have the image look like it was taken during a power outage} & \itext{Dim the lighting in the scene} & \itext{Replace the background with darkness} & \itext{Add a candle on the table}  \\
         & \lines~ \\
         \multicolumn{6}{c}{\vspace{-20pt}}\\
         
         \midrule

          \raisebox{-10pt}{\rotatebox{90}{\textit{HQEdit}}}
         & \includegraphics[width=0.19\textwidth, height=0.19\textwidth]{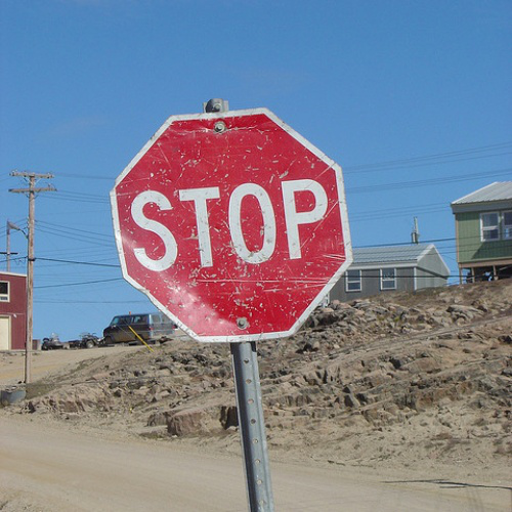} & 
         \includegraphics[width=0.19\textwidth, height=0.19\textwidth]{figures/qual_results/000128_orig.png} & 
         \includegraphics[width=0.19\textwidth, height=0.19\textwidth]{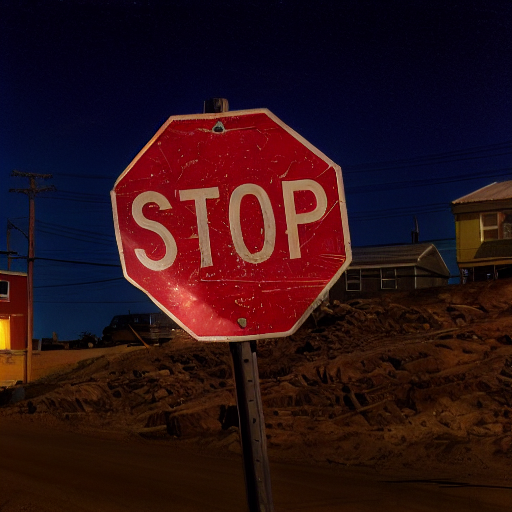} & 
         \includegraphics[width=0.19\textwidth, height=0.19\textwidth]{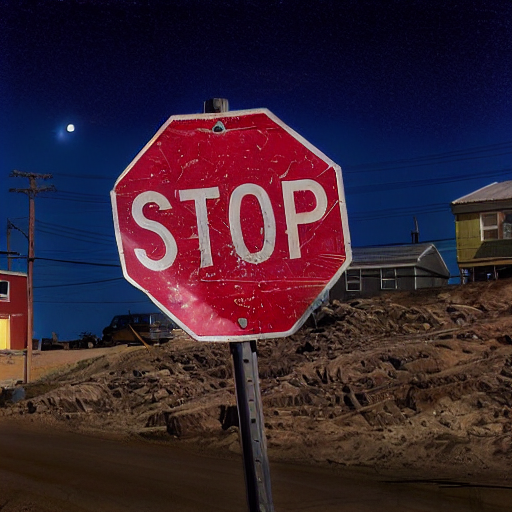} & 
         \includegraphics[width=0.19\textwidth, height=0.19\textwidth]{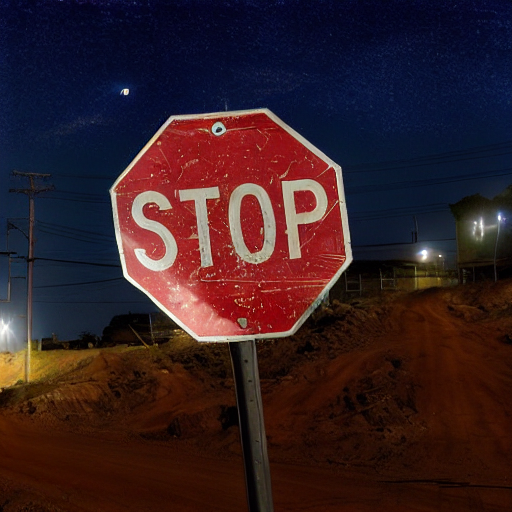} \\
         & Instructions~$\rightarrow$ & \itext{Change the time of the day to night} & \itext{Replace the bright sky with a starry night sky} & \itext{Add a moon in the background} & \itext{Add streetlights along the dirt road}  \\
         & \lines~ \\
         \multicolumn{6}{c}{\vspace{-20pt}}\\
         
         \midrule

         \end{tabularx}
    }
    \caption{\textbf{Additional qualitative results for InstructPix2Pix.}}\label{fig:supp-qual-1}
\end{figure*}

\begin{figure*}[t]
    \centering
    \setlength{\tabcolsep}{0pt} %
    \resizebox{\linewidth}{!}{
    \begin{tabularx}{\textwidth}{@{}H@{\hspace{0.01\textwidth}}Y@{\hspace{0.01\textwidth}}Y@{\hspace{0.01\textwidth}}YYY@{}}
    & & & \multicolumn{3}{c}{\methodname}\\
    & & Baseline & \small{$N=1$} & \small{$N=2$} & \small{$N=3$}\\
    \cmidrule(lr){3-3}\cmidrule(lr){4-6}
         & Original image & \sqHeader~ \\\midrule
         \raisebox{-5pt}{\rotatebox{90}{\textit{IP2P}}}
         & \includegraphics[width=0.19\textwidth, height=0.19\textwidth]{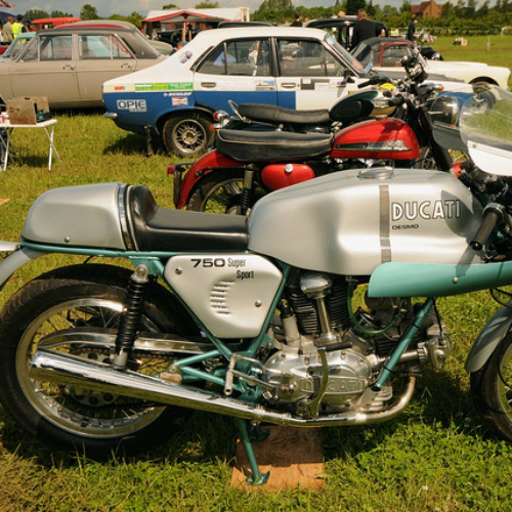} & 
         \includegraphics[width=0.19\textwidth, height=0.19\textwidth]{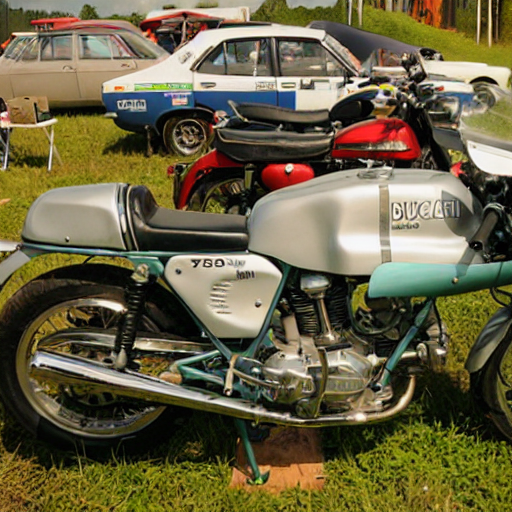} & 
         \includegraphics[width=0.19\textwidth, height=0.19\textwidth]{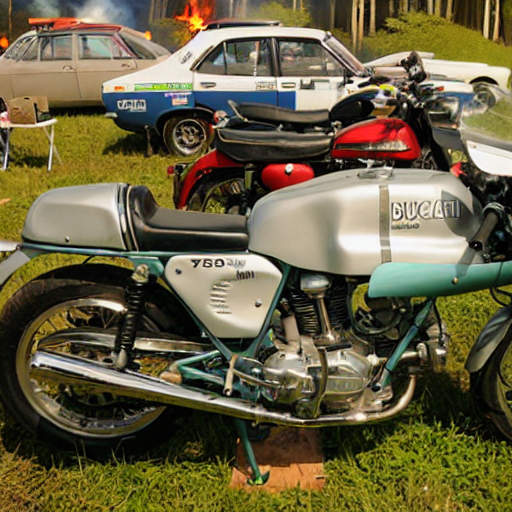} & 
         \includegraphics[width=0.19\textwidth, height=0.19\textwidth]{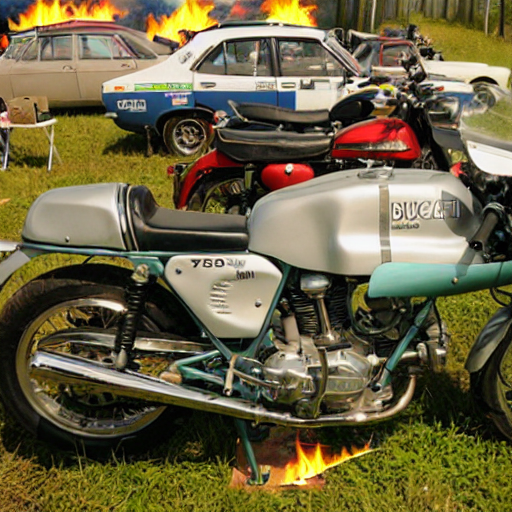} & 
         \includegraphics[width=0.19\textwidth, height=0.19\textwidth]{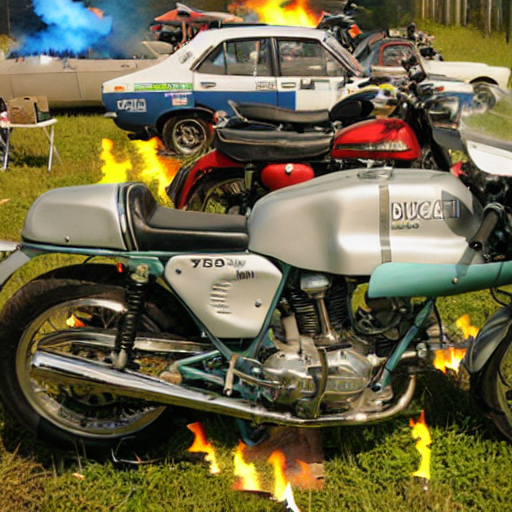} \\
         & Instructions~$\rightarrow$ & \itext{Change the image to appear like a fire in a forest} & \itext{Replace the grassy field with a burning forest background} & \itext{Add flames around the Ducati motorcycle} & \itext{Add smoke in the sky above the forest}  \\
         & \lines~ \\
         \multicolumn{6}{c}{\vspace{-20pt}}\\
         \midrule
         \raisebox{-5pt}{\rotatebox{90}{\textit{IP2P}}}
         & \includegraphics[width=0.19\textwidth, height=0.19\textwidth]{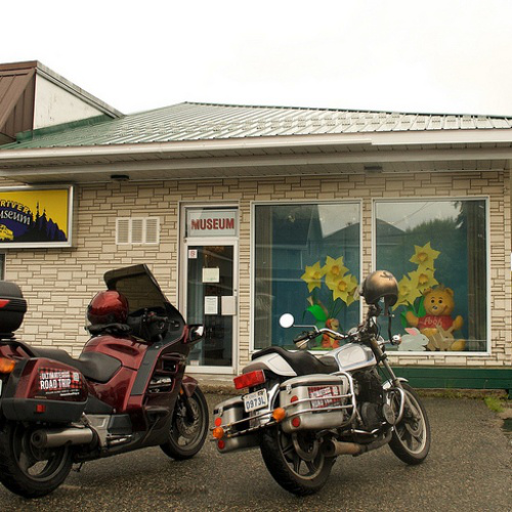} & 
         \includegraphics[width=0.19\textwidth, height=0.19\textwidth]{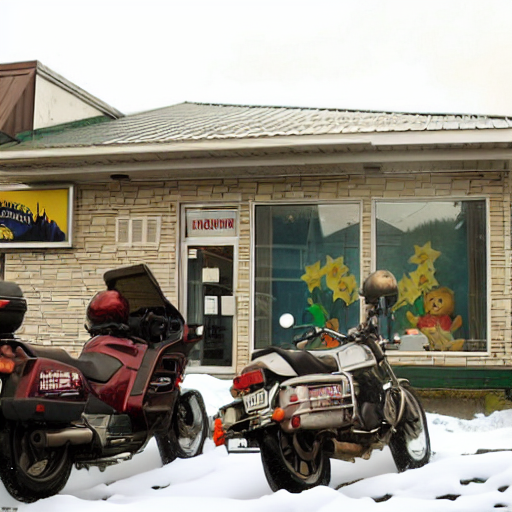} & 
         \includegraphics[width=0.19\textwidth, height=0.19\textwidth]{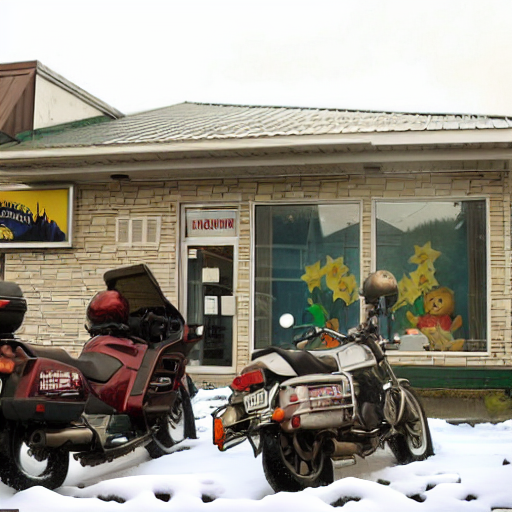} & 
         \includegraphics[width=0.19\textwidth, height=0.19\textwidth]{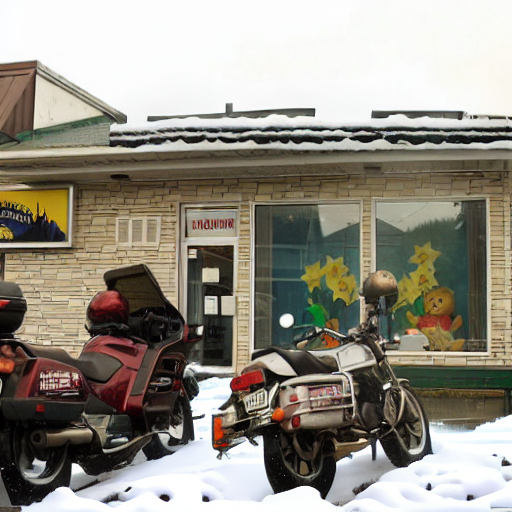} & 
         \includegraphics[width=0.19\textwidth, height=0.19\textwidth]{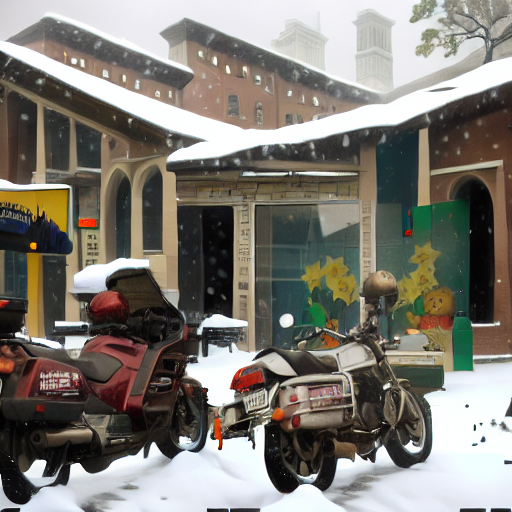} \\
         & Instructions~$\rightarrow$ & \itext{Make it a snowy day}  & \itext{Add snow on the ground around the motorcycles} & \itext{Add snow to the roofs and ledges of the museum building} & \itext{Add falling snowflakes throughout the scene} \\
         & \lines~ \\
         \multicolumn{6}{c}{\vspace{-20pt}}\\
         \midrule
         \raisebox{-5pt}{\rotatebox{90}{\textit{IP2P}}}
         & \includegraphics[width=0.19\textwidth, height=0.19\textwidth]{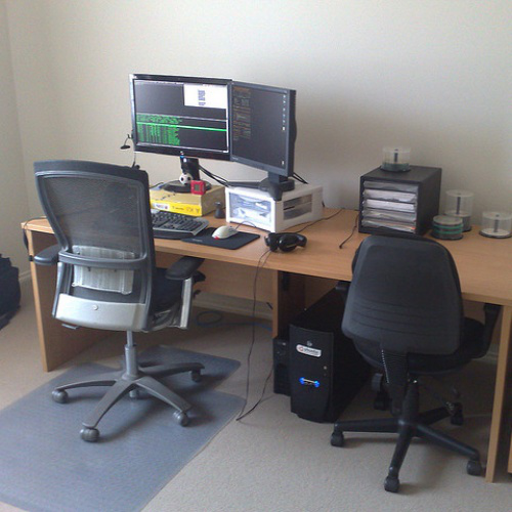} & 
         \includegraphics[width=0.19\textwidth, height=0.19\textwidth]{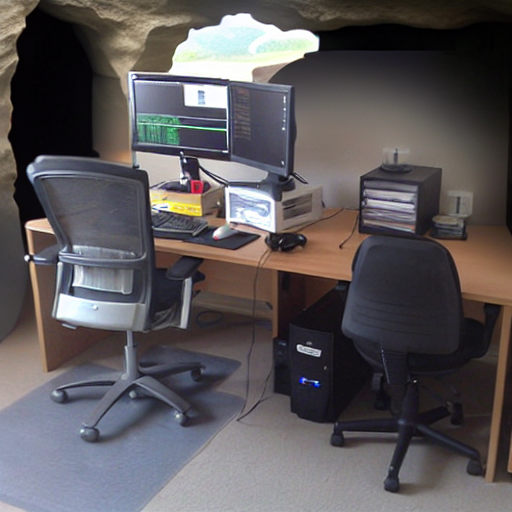} & 
         \includegraphics[width=0.19\textwidth, height=0.19\textwidth]{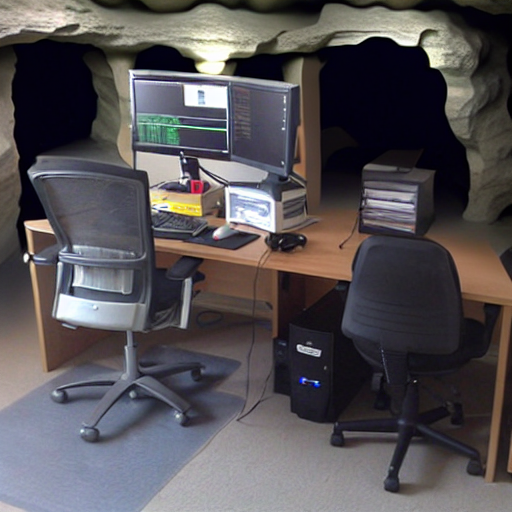} & 
         \includegraphics[width=0.19\textwidth, height=0.19\textwidth]{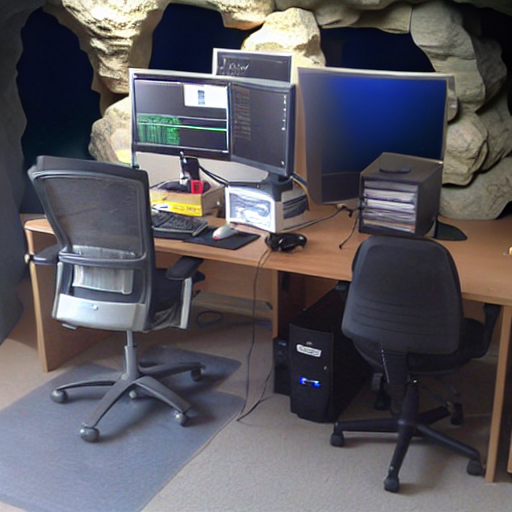} & 
         \includegraphics[width=0.19\textwidth, height=0.19\textwidth]{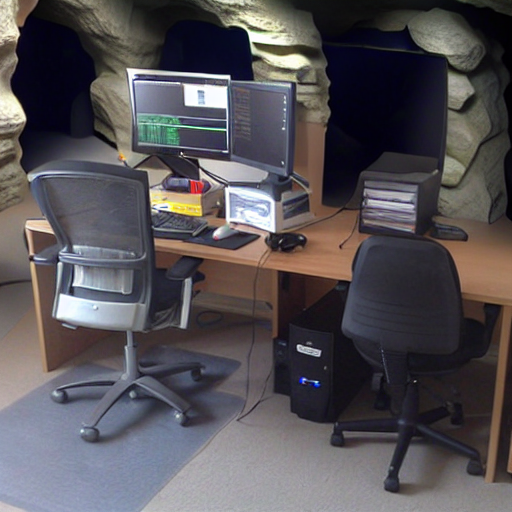} \\
         & Instructions~$\rightarrow$ & \itext{Put this inside of a cave}  & \itext{Replace the background with a cave interior} & \itext{Add rocky textures to the monitor, keyboard, mouse, and speakers} & \itext{Adjust the lighting to darker, more cave-like conditions} \\
         & \lines~ \\
         \multicolumn{6}{c}{\vspace{-20pt}}\\
         \midrule
         \raisebox{-5pt}{\rotatebox{90}{\textit{IP2P}}}
         & \includegraphics[width=0.19\textwidth, height=0.19\textwidth]{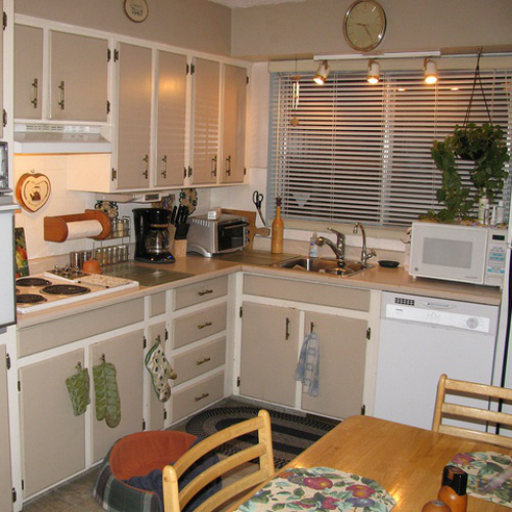} & 
         \includegraphics[width=0.19\textwidth, height=0.19\textwidth]{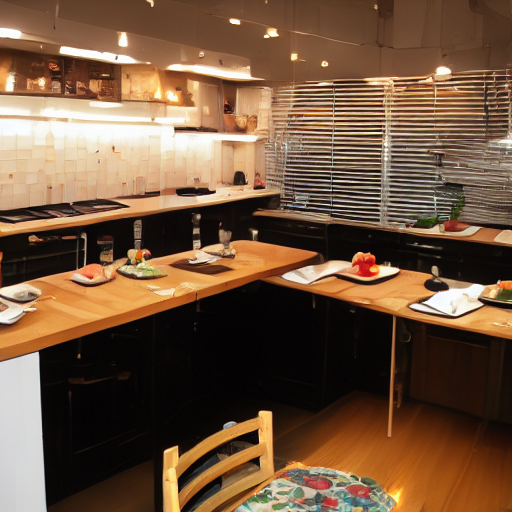} & 
         \includegraphics[width=0.19\textwidth, height=0.19\textwidth]{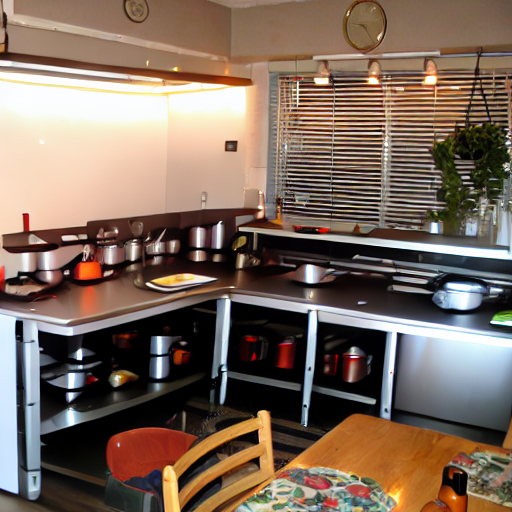} & 
         \includegraphics[width=0.19\textwidth, height=0.19\textwidth]{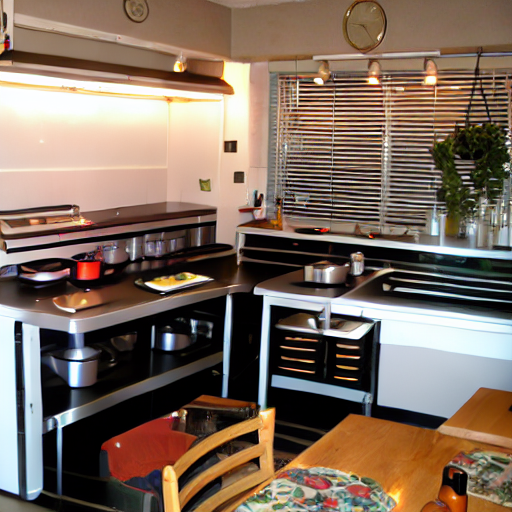} & 
         \includegraphics[width=0.19\textwidth, height=0.19\textwidth]{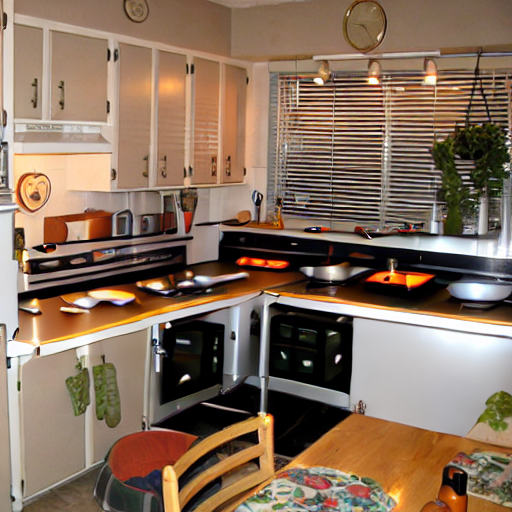} \\
         & Instructions~$\rightarrow$ & \itext{Change this to a restaurant kitchen}  & \itext{Replace the kitchen table with an industrial stainless-steel worktable} & \itext{Add commercial-grade ovens and stoves} & \itext{Add multiple professional kitchen knives and utensils on the countertop} \\
         & \lines~ \\
         \multicolumn{6}{c}{\vspace{-20pt}}\\
         \midrule
         \end{tabularx}
    }
    \caption{\textbf{Qualitative results for MagicBrush.}}\label{fig:supp-qual-2}
\end{figure*}

\begin{figure*}[t]
    \centering
    \setlength{\tabcolsep}{0pt} %
    \resizebox{\linewidth}{!}{
    \begin{tabularx}{\textwidth}{@{}H@{\hspace{0.01\textwidth}}Y@{\hspace{0.01\textwidth}}Y@{\hspace{0.01\textwidth}}YYY@{}}
    & & & \multicolumn{3}{c}{\methodname}\\
    & & Baseline & \small{$N=1$} & \small{$N=2$} & \small{$N=3$}\\
    \cmidrule(lr){3-3}\cmidrule(lr){4-6}
         & Original image & \sqHeader~ \\\midrule
         \raisebox{-20pt}{\rotatebox{90}{\textit{MagicBrush}}}
         & \includegraphics[width=0.19\textwidth, height=0.19\textwidth]{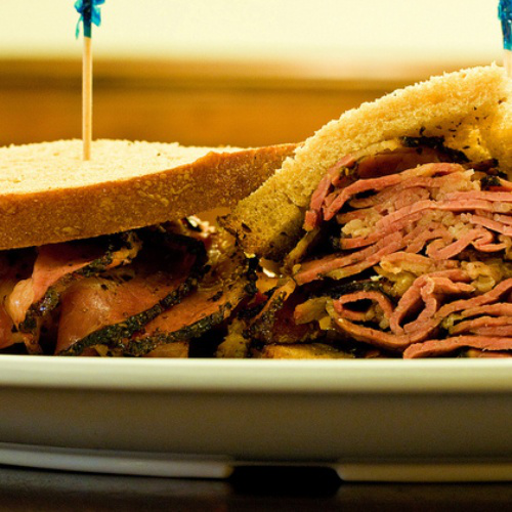} & 
         \includegraphics[width=0.19\textwidth, height=0.19\textwidth]{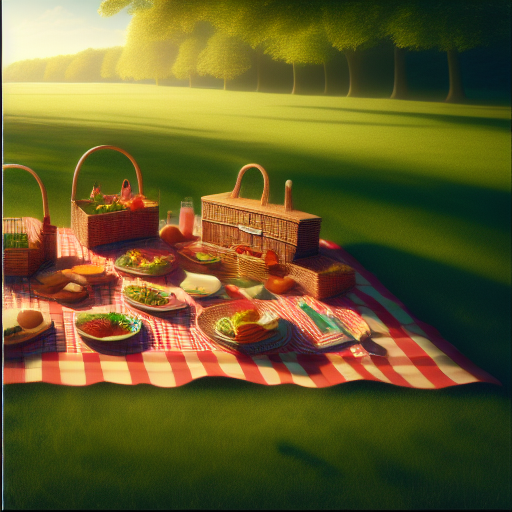} & 
         \includegraphics[width=0.19\textwidth, height=0.19\textwidth]{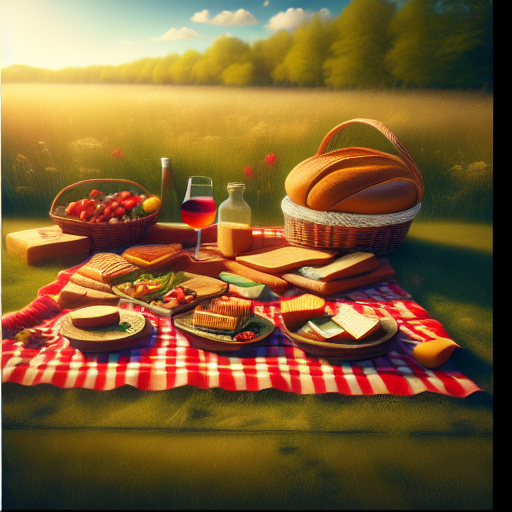} & 
         \includegraphics[width=0.19\textwidth, height=0.19\textwidth]{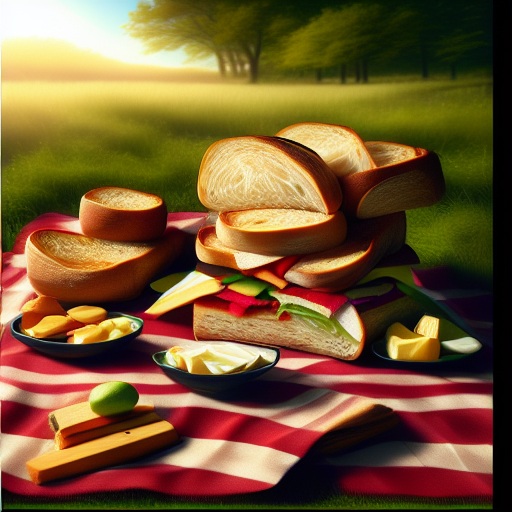} & 
         \includegraphics[width=0.19\textwidth, height=0.19\textwidth]{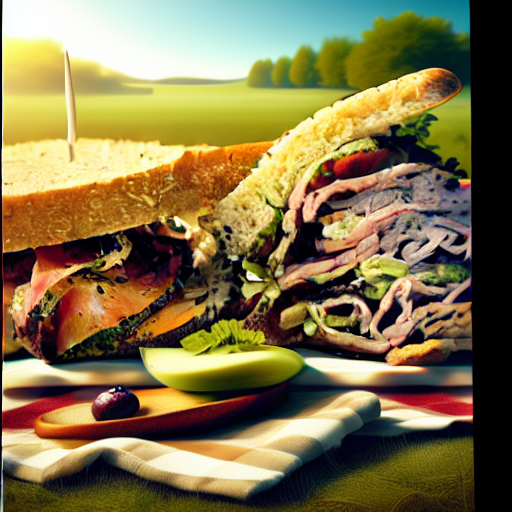} \\
         & Instructions~$\rightarrow$ & \itext{Make the scene seem like it is an outdoor picnic} & \itext{Add a picnic blanket under the sandwich} & \itext{Add trees in the background} & \itext{Add a basket with fruits beside the sandwich}  \\
         & \lines~ \\
         \multicolumn{6}{c}{\vspace{-20pt}}\\
         
         \midrule
         \raisebox{-20pt}{\rotatebox{90}{\textit{MagicBrush}}}
         & \includegraphics[width=0.19\textwidth, height=0.19\textwidth]{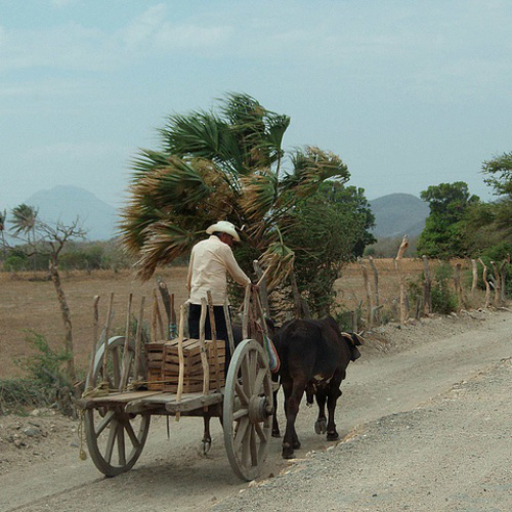} & 
         \includegraphics[width=0.19\textwidth, height=0.19\textwidth]{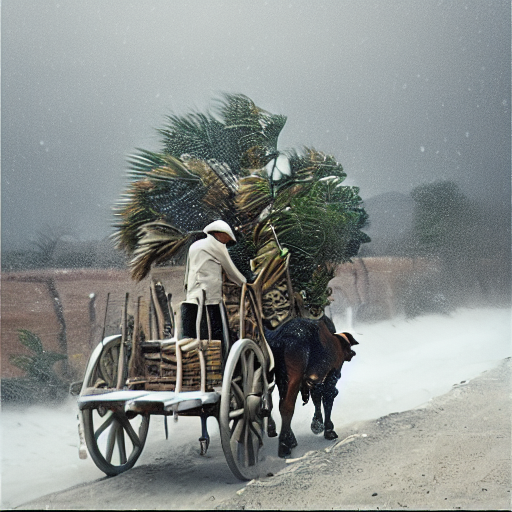} & 
         \includegraphics[width=0.19\textwidth, height=0.19\textwidth]{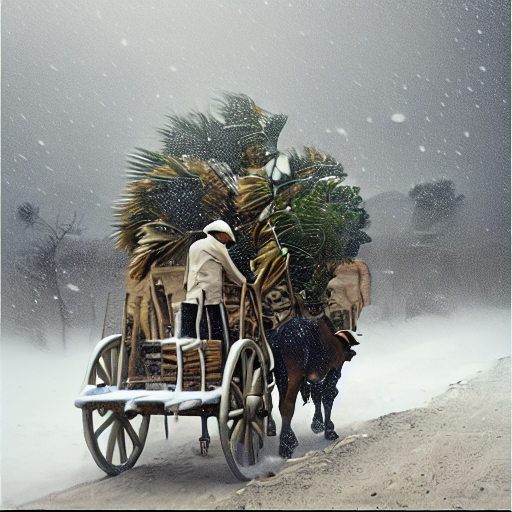} & 
         \includegraphics[width=0.19\textwidth, height=0.19\textwidth]{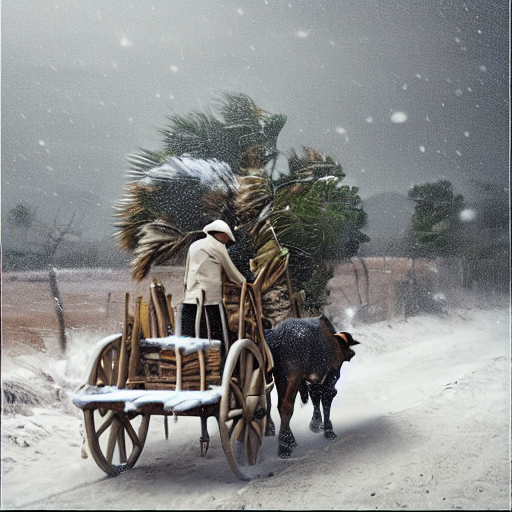} & 
         \includegraphics[width=0.19\textwidth, height=0.19\textwidth]{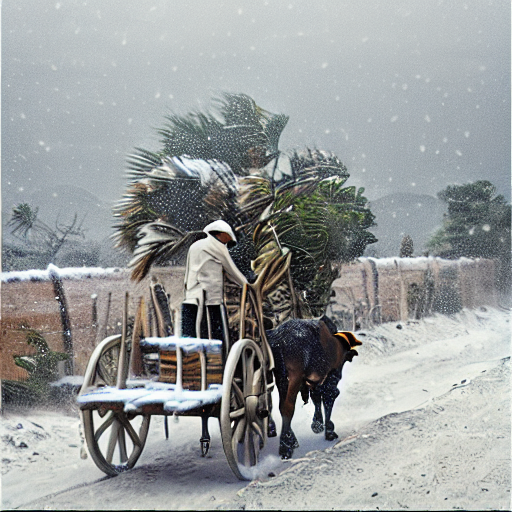} \\
         & Instructions~$\rightarrow$ & \itext{Change this to a snowstorm} & \itext{Add falling snow to the scene} & \itext{Add snow accumulation to the cart and cow} & \itext{Replace the clear sky with a cloudy, snow-filled sky}  \\
         & \lines~ \\
         \multicolumn{6}{c}{\vspace{-20pt}}\\
         
         \midrule

         \raisebox{-20pt}{\rotatebox{90}{\textit{MagicBrush}}}
         & \includegraphics[width=0.19\textwidth, height=0.19\textwidth]{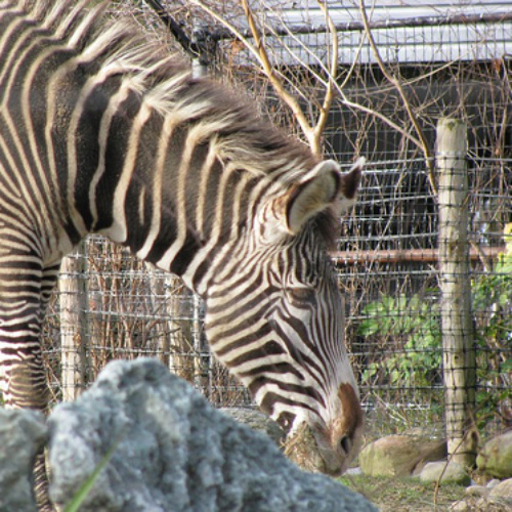} & 
         \includegraphics[width=0.19\textwidth, height=0.19\textwidth]{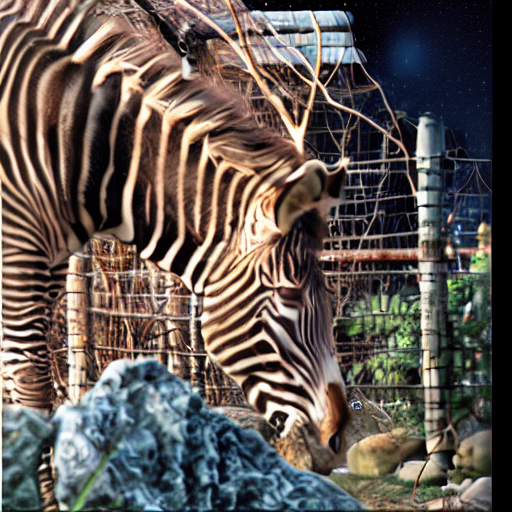} & 
         \includegraphics[width=0.19\textwidth, height=0.19\textwidth]{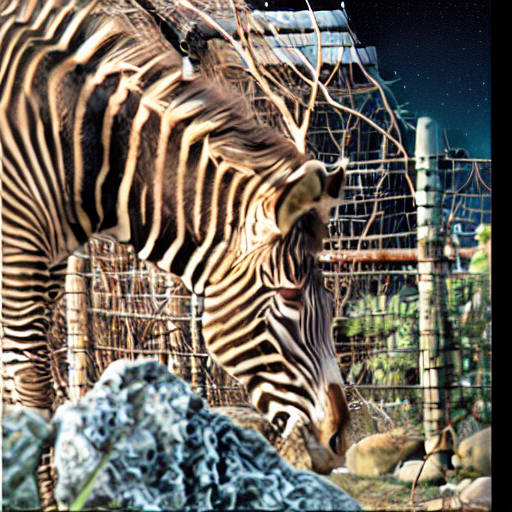} & 
         \includegraphics[width=0.19\textwidth, height=0.19\textwidth]{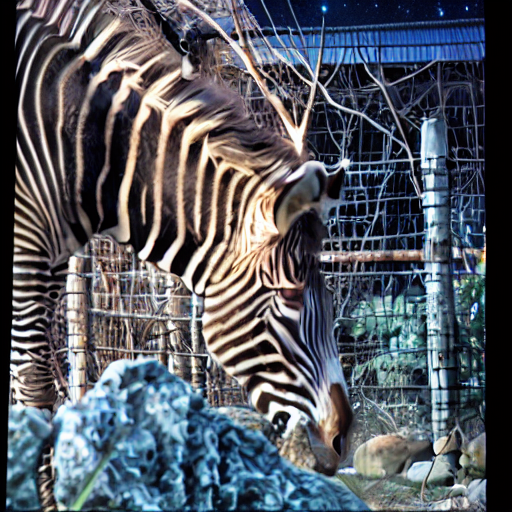} & 
         \includegraphics[width=0.19\textwidth, height=0.19\textwidth]{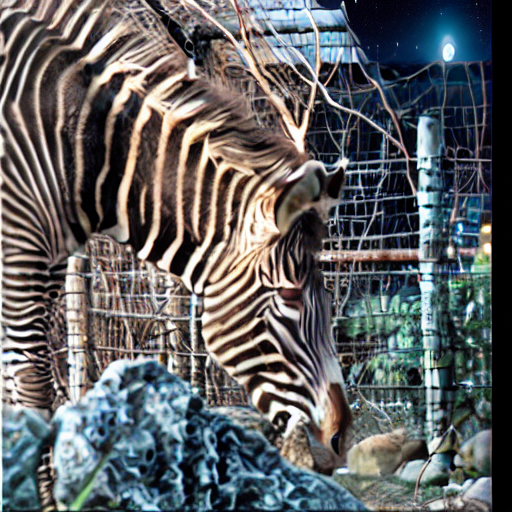} \\
         & Instructions~$\rightarrow$ & \itext{Change this to night time with lots of stars} & \itext{Change the sky to a dark, star-filled sky} & \itext{Adjust the lighting to create moonlit shadows} & \itext{Add a glowing moon in the background}  \\
         & \lines~ \\
         \multicolumn{6}{c}{\vspace{-20pt}}\\
         
         \midrule

         \raisebox{-20pt}{\rotatebox{90}{\textit{MagicBrush}}}
         & \includegraphics[width=0.19\textwidth, height=0.19\textwidth]{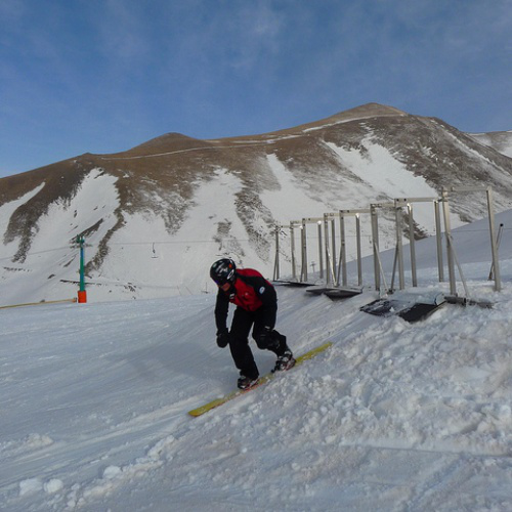} & 
         \includegraphics[width=0.19\textwidth, height=0.19\textwidth]{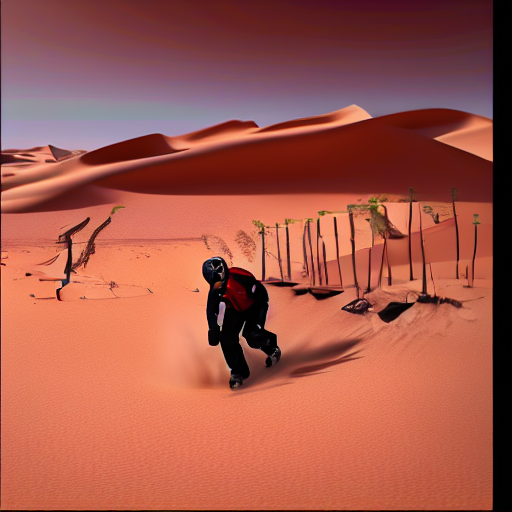} & 
         \includegraphics[width=0.19\textwidth, height=0.19\textwidth]{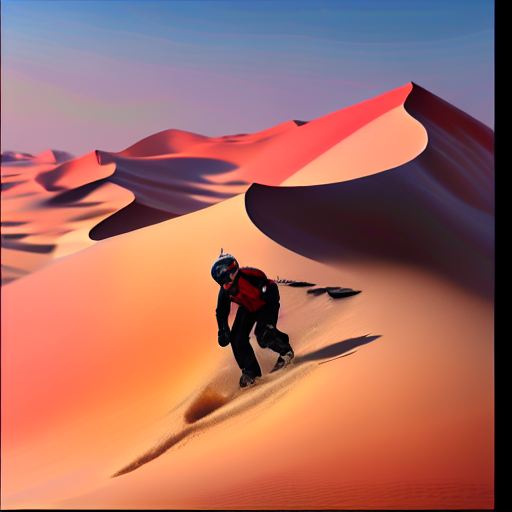} & 
         \includegraphics[width=0.19\textwidth, height=0.19\textwidth]{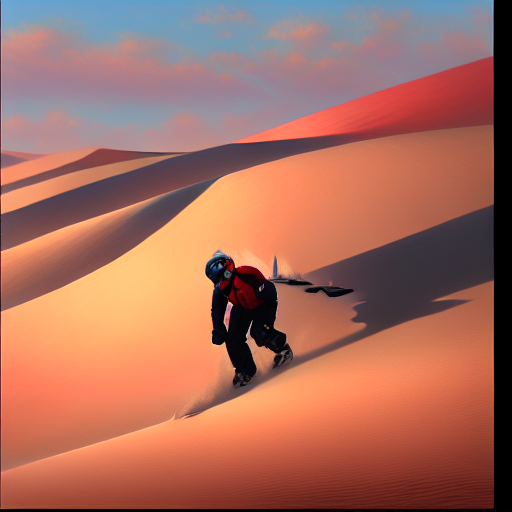} & 
         \includegraphics[width=0.19\textwidth, height=0.19\textwidth]{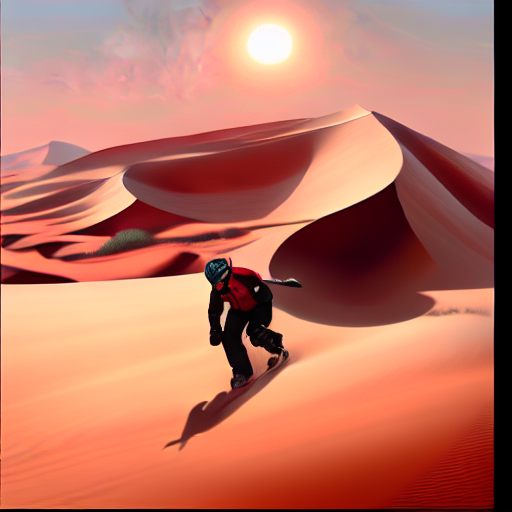} \\
         & Instructions~$\rightarrow$ & \itext{Make the image appear to be in the Sahara desert} & \itext{Replace the snowy hill with sand dunes} & \itext{Change the man's red jacket to a lighter, desert-appropriate color} & \itext{Add a blazing sun in the sky}  \\
         & \lines~ \\
         \multicolumn{6}{c}{\vspace{-20pt}}\\
         
         \midrule
         \end{tabularx}
    }
    \caption{\textbf{Qualitative results for HQEdit.}}\label{fig:supp-qual-3}
\end{figure*}

\paragraph{Additional qualitative results}
We provide additional qualitative results for \methodname applied on top of all three baselines. In particular, in Figure~\ref{fig:supp-qual-1} we show more editing examples using InstructPix2Pix as a baseline, complementing qualitative results in the main paper (Figure~\ref{fig:qual}), while in Figure~\ref{fig:supp-qual-2} and Figure~\ref{fig:supp-qual-3} we report results on MagicBrush and HQEdit baselines, respectively, which are not included in the main manuscript due to limited space.

\end{document}